\documentclass[default,iicol]{sn-jnl}

\jyear{2022}%

\theoremstyle{thmstyleone}%
\theoremstyle{thmstyletwo}

\theoremstyle{thmstylethree}

\usepackage{overpic}
\usepackage{color}
\usepackage{colortbl}
\usepackage{marvosym}

\newcommand{\figref}[1]{Fig.~\ref{#1}}
\newcommand{\tabref}[1]{Table~\ref{#1}}
\newcommand{\secref}[1]{Sec.~\ref{#1}}

\def\ie{\emph{i.e.}}
\def\eg{\emph{e.g.}}

\def\etal{{\em et al.}}

\definecolor{mygray}{gray}{.95}
\definecolor{myRed}{RGB}{219, 68, 55}
\definecolor{myGreen}{RGB}{15, 157, 88}
\definecolor{myBlue}{RGB}{66, 133, 244}
\newcommand{\tr}[1]{{\textcolor{myRed}{\textbf{#1}}}}
\newcommand{\tg}[1]{{\textcolor{myGreen}{\textbf{#1}}}}
\newcommand{\tb}[1]{{\textcolor{myBlue}{\textbf{#1}}}}
\newcommand{\jgp}[1]{{\textcolor{black}{#1}}}

\newcommand{\res}[1]{{\textcolor{black}{#1}}}   

\newcommand{\equref}[1]{Equ.(\ref{#1})}
\newcommand{\tabincell}[2]{\begin{tabular}{@{}#1@{}}#2\end{tabular}}

\def\ourmodel{DGNet}
\def\ourmodelS{DGNet-S}
\def\ourmodelL{DGNet}

\newcommand\blfootnote[1]{
\begingroup
\renewcommand\thefootnote{}\footnote{#1}
\addtocounter{footnote}{-1}
\endgroup
}

\begin{document}

\title[DGNet]{\textbf{Deep Gradient Learning for Efficient Camouflaged Object Detection}}

\author[1]{\fnm{Ge-Peng} \sur{Ji}}

\author[2]{\fnm{Deng-Ping} \sur{Fan}$\textsuperscript{\Letter}$}

\author[1]{\fnm{Yu-Cheng} \sur{Chou}}

\author[3]{\fnm{Dengxin} \sur{Dai}}

\author[2]{\fnm{Alexander} \sur{Liniger}}

\author[2]{\fnm{Luc Van} \sur{Gool}}

\affil[1]{\orgdiv{School of Computer Science}, \orgname{Wuhan University}, \orgaddress{\city{Wuhan}, \country{China}}}

\affil[2]{\orgdiv{Computer Vision Lab}, \orgname{ETH Zürich}, \orgaddress{\city{Zürich}, \country{Switzerland}}}

\affil[3]{\orgdiv{Vision for Autonomous Systems Group}, \orgname{MPI for Informatics}, \orgaddress{\city{Saarbrücken}, \country{Germany}}}

\abstract{
This paper introduces \textbf{\ourmodel}, a novel deep framework that exploits object gradient supervision for camouflaged object detection (COD).
It decouples the task into two connected branches, \ie, a context and a texture encoder. The essential connection is the gradient-induced transition, representing a soft grouping between context and texture features. 
Benefiting from the simple but efficient framework, \ourmodel~outperforms existing state-of-the-art COD models by a large margin.
Notably, our efficient version, \textbf{\ourmodelS}, runs in real-time (80 fps) and achieves comparable results to the cutting-edge model JCSOD-CVPR$_{21}$ with only 6.82\% parameters. Application results also show that the proposed \ourmodel~performs well in polyp segmentation, defect detection, and transparent object segmentation tasks.
Codes will be made available at \href{https://github.com/GewelsJI/DGNet}{https://github.com/GewelsJI/DGNet}.
}


\maketitle

\section{Introduction}

Camouflaged object detection~\cite{fan2020camouflaged,fan2021concealed} (COD) aims to segment objects with either artificial or natural patterns where objects `perfectly' blend into the background to avoid being discovered~\cite{fan2021concealed}.
Several successful applications, such as medical image analysis (\eg, polyp~\cite{fan2020pranet,ji2021pnsnet,ji2022vps} and lung infection~\cite{fan2020inf,wu2021jcs,liu2021covid} segmentation), \res{video understanding (\eg, surveillance~\cite{chen2022pedestrian} and autonomous driving~\cite{xue2018survey})} and recreational art~\cite{feng2013facilitating,dean2017art}, have shown COD's scientific and practical value. \blfootnote{$\textsuperscript{\Letter}$ Corresponding author (dengpfan@gmail.com). The major part of this work was done while Ge-Peng Ji was an intern mentored by Deng-Ping Fan.}

\begin{figure}[t!]
    \centering
    \begin{overpic}[width=\linewidth]{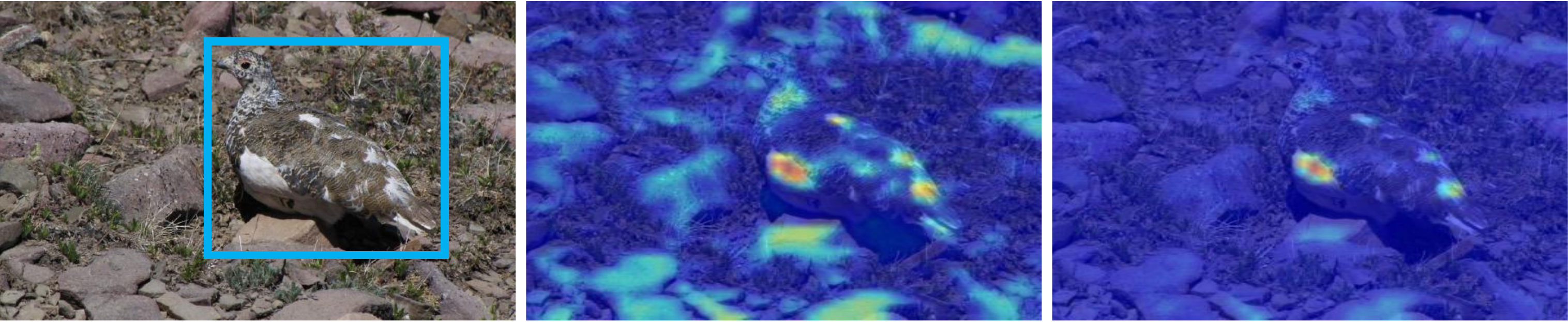}
    \put(0, -4){\footnotesize Camouflaged Image}
    \put(32.7, -4){\footnotesize (a) Object Boundary}
    \put(66.8, -4){\footnotesize (b) Object Gradient}
    \end{overpic}
    \vspace{0.2pt}
    \caption{Feature visualization of learned texture. We observe that the proposed \ourmodelS~under the object boundary supervision (a) contains diffused noises in the background. By contrast, object gradient supervision (b) enforces the network focus on the regions where the intensity changes dramatically.}
    \label{fig:feature_vis}
\end{figure}

Recent studies~\cite{mei2021camouflaged,lyu2021simultaneously,jia2022segment,zhong2022detecting} present compelling results based on the supervision of the \jgp{\textit{whole object-level}} ground-truth mask. Later, various sophisticated techniques, \eg, boundary-based~\cite{zhai2021mutual,ji2021fast,zhu2022can} and uncertainty-guided~\cite{yang2021uncertainty,li2021uncertainty}, were developed to augment COD's underlying representations.
However, features learned from boundary-supervised or uncertainty-based models usually respond to the \jgp{sparse} edge of camouflage objects, thereby introducing noisy features, especially for complex scenes (see \figref{fig:feature_vis}-a). Besides, the boundaries of camouflaged objects are always `indefinable' or `fuzzy'; 
thus, they do not be pop-out from a quick visual scanning.
We notice that despite the object's camouflage, there are still some clues left, shown in the first column of \figref{fig:feature_vis} (white speckles). Instead of extracting only boundary or uncertainty regions, we are interested in how the network mines these `discriminative patterns' inside the object.

From this perspective, we present our \textit{deep gradient network} (\textbf{DGNet})~via the explicit supervision of the \textit{object-level} gradient map. The underlying hypothesis is that some intensity changes inside the camouflaged objects.  
To ease the learning task, we decouple the \ourmodel~into two connected branches, \ie, a context and a texture encoder. 
The former can be viewed as a contextual semantics learner, while the latter acts as a structural texture extractor. In this way, we can alleviate the feature ambiguity between the high-level and low-level features extracted from the individual branch.
To sufficiently aggregate the above two discriminative features generated by the two branches, we further design a gradient-induced transition (GIT) module that collaboratively ensembles the multi-source feature space at different group scales (\ie, soft grouping).
\figref{fig:feature_vis}-b shows that our \ourmodel~can detect texture patterns while suppressing the background noise by an intensity-sensitive strategy focusing on the intra-region of a camouflaged object.

Extensive experiments on three challenging COD benchmarks illustrate that the proposed \ourmodel~achieves state-of-the-art (SOTA) performance without introducing any complicated structures. 
Furthermore, we implement an efficient version, \ourmodelS, with 8.3M parameters, which achieves the fastest inference speed (80 fps) among COD-related baselines.
Notably, it only has 6.82\% parameters compared to the cutting-edge model JCSOD-CVPR$_{21}$~\cite{li2021uncertainty} while achieving comparable performance.
These results show that \ourmodel~significantly narrows the gap between scientific research and practical application. Three downstream applications (see~\secref{sec:apps}) of our \ourmodel~also support this conclusion. 
The major \textbf{contributions} of this paper are summarized as follows: 
\begin{itemize}
\item[$\bullet$] We introduce a novel deep gradient-based framework, dubbed \textbf{\ourmodel}, for addressing the camouflaged object detection task.

\item[$\bullet$] We propose a \textbf{gradient-induced transition} to automatically group features from the context and texture branches according to \jgp{the soft grouping strategy}.

\item[$\bullet$] We present \textbf{three applications} and achieve good performance, including polyp segmentation, defect detection, and transparent object segmentation.

\end{itemize}

\section{Prior Works}
\noindent\textbf{Traditional} methods detect camouflaged objects via extracting various hand-crafted features between the camouflaged areas and their backgrounds, which calculate the 3D convexity~\cite{pan2011study}, co-occurrence matrix~\cite{sengottuvelan2008performance}, expectation-maximization statistics~\cite{liu2012foreground}, optical flow~\cite{hou2011detection}, and Gaussian mixture model~\cite{gallego2014foreground}.
Those methods work well for simple backgrounds, while the performance degrades drastically for complex backgrounds.

\noindent\textbf{CNN-based} approaches could be generally categorized into three strategies:
\textit{a) Attention-based strategy:}
Sun~\etal~\cite{sun2021Context,chen2022camouflaged} introduce a network with an attention-induced cross-level fusion module to integrate multi-scale features and a dual-branch global context module to mine multi-scale contextual information.
To mimic the detection process of predators, Mei~\etal~\cite{mei2021camouflaged} develop PFNet, which contains a positioning and focusing module to conduct the identification.
\cite{ren2021deep,ji2021fast} propose delicate structures such as covariance matrices of feature and multivariate calibration components to improve the robustness of the network.
Kajiura~\etal~\cite{Kajiura2021improving} improve the detection accuracy via exploring the uncertainties of pseudo-edge and pseudo-map labels.
Zhuge~\etal~\cite{zgmc2022cubenet} propose a cube-alike architecture for COD, which accompanies attention fusion and X-shaped connection to integrate multiple-layer features \res{sufficiently}.
\textit{b) Two-stage strategy:}
%
Search and identification strategy~\cite{fan2020camouflaged} is an early practice to model the COD task. In~\cite{fan2021concealed}, the neighbour connection decoder and group-reversal attention are introduced in SINet~\cite{fan2020camouflaged} to boost the performance further.
\textit{c) Joint-learning strategy:}
ANet~\cite{le2019anabranch} is an early attempt to utilize the classification and segmentation scheme for COD.
LSR~\cite{lyu2021simultaneously} and JCSOD~\cite{li2021uncertainty} have recently renewed the joint-learning framework by introducing camouflaged ranking or learning from salient objects to camouflaged objects.
ZoomNet~\cite{pang2022zoom} is a mixed-scale triplet network that employs the zoom strategy to learn the discriminative camouflaged semantics.

\noindent\textbf{Transformer-based \& Graph-based} models are two recent technology trends. 
Recently, Mao~\etal~\cite{mao2021transformer} introduce the concept of difficulty-aware learning based on the Transformer for both camouflaged and salient object detection.
UGTR~\cite{yang2021uncertainty} explicitly utilizes the probabilistic representational model to learn the uncertainties of the camouflaged object under the Transformer framework. 
In addition, Cheng~\etal~\cite{cheng2022implicit} are the first to collect a video dataset for COD and utilise the Transformer-based framework to exploit short-term dynamics and long-term temporal consistency to detect dynamic camouflaged objects.
%
Later, Zhai~\etal~\cite{zhai2021mutual} design the mutual graph learning model, which decouples one input into different features for roughly locating the target and accurately capturing its boundary.

\begin{figure}[t!]
    \centering
    \begin{overpic}[width=\linewidth]{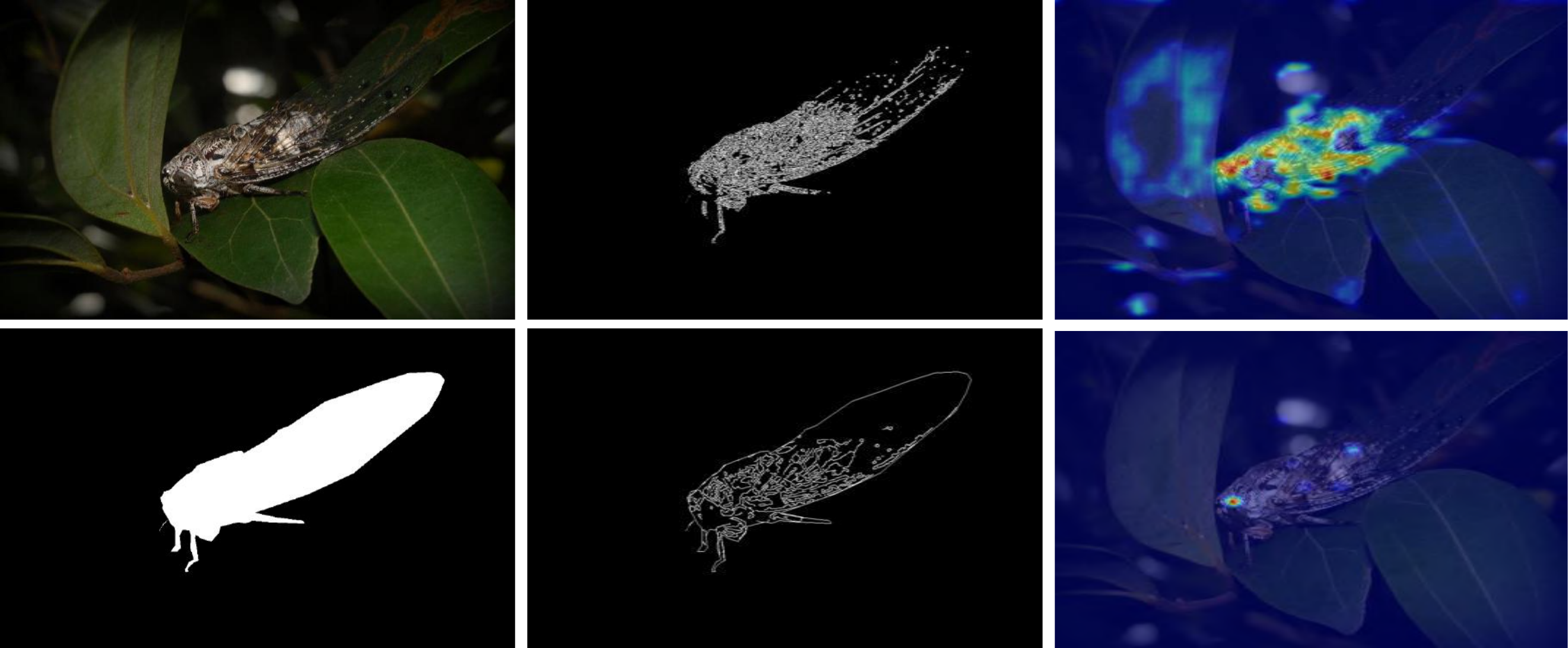}
    \put(0.5, 22.5){\footnotesize \color{white} Camouflaged Image}
    \put(37, 22.5){\footnotesize \color{white} (a) \ourmodel-Grad}
    \put(55, 38){\footnotesize \color{white} \textit{\textbf{(Ours)}}}
    \put(74, 22.5){\footnotesize \color{white} (b) Feature}
    \put(88.5, 38){\footnotesize \color{white} \textit{\textbf{(Ours)}}}
    \put(6, 1.5){\footnotesize \color{white} Ground-truth}
    \put(38, 1.5){\footnotesize \color{white} (c) TINet-Text}
    \put(74, 1.5){\footnotesize \color{white} (d) Feature}
    \end{overpic}
    \caption{\res{Compared to} the texture label proposed in TINet~\cite{zhu2021inferring}, our object gradient label (a) keeps more geometric cues inside the camouflaged object. \ourmodel~under the supervision of texture label (c) fails to infer attentive regions (d) since the imbalanced distribution of sparse pixels (\eg, thin object boundaries). Notably, such improvement exerts our \ourmodel~more robust with the reliable auxiliaries, \eg, feature in (b).}
    \label{fig:grad_comp}
\end{figure}
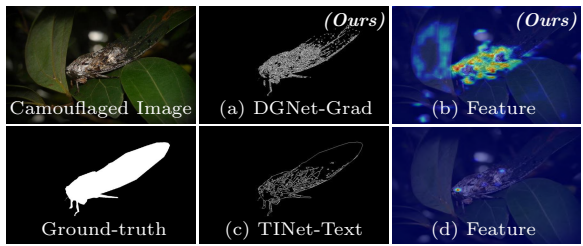

\noindent\textbf{Remarks.} By contrast, our work excavates the texture information via learning the object-level gradient rather than using boundary-aware or uncertainty-aware modelling. The biologically inspired idea behind this is that the abundant gradient cues inside the camouflaged object deserve to be explored, while the sparse boundary cues are insufficient to achieve this. As shown in~\figref{fig:grad_comp}, we also note that the recent work~\cite{zhu2021inferring} tries to utilize the texture cues while they discard excessive object gradient cues due to different threshold settings of the Canny detector. In short, this paper aims to design an elegant framework towards efficient COD with more concise ideas (\ie, object gradient learning). More experimental validations are discussed in~\secref{sec:ablation}.

\section{Deep Gradient Network}
\begin{figure*}[t!]
    \centering
    \begin{overpic}[width=0.84\linewidth]{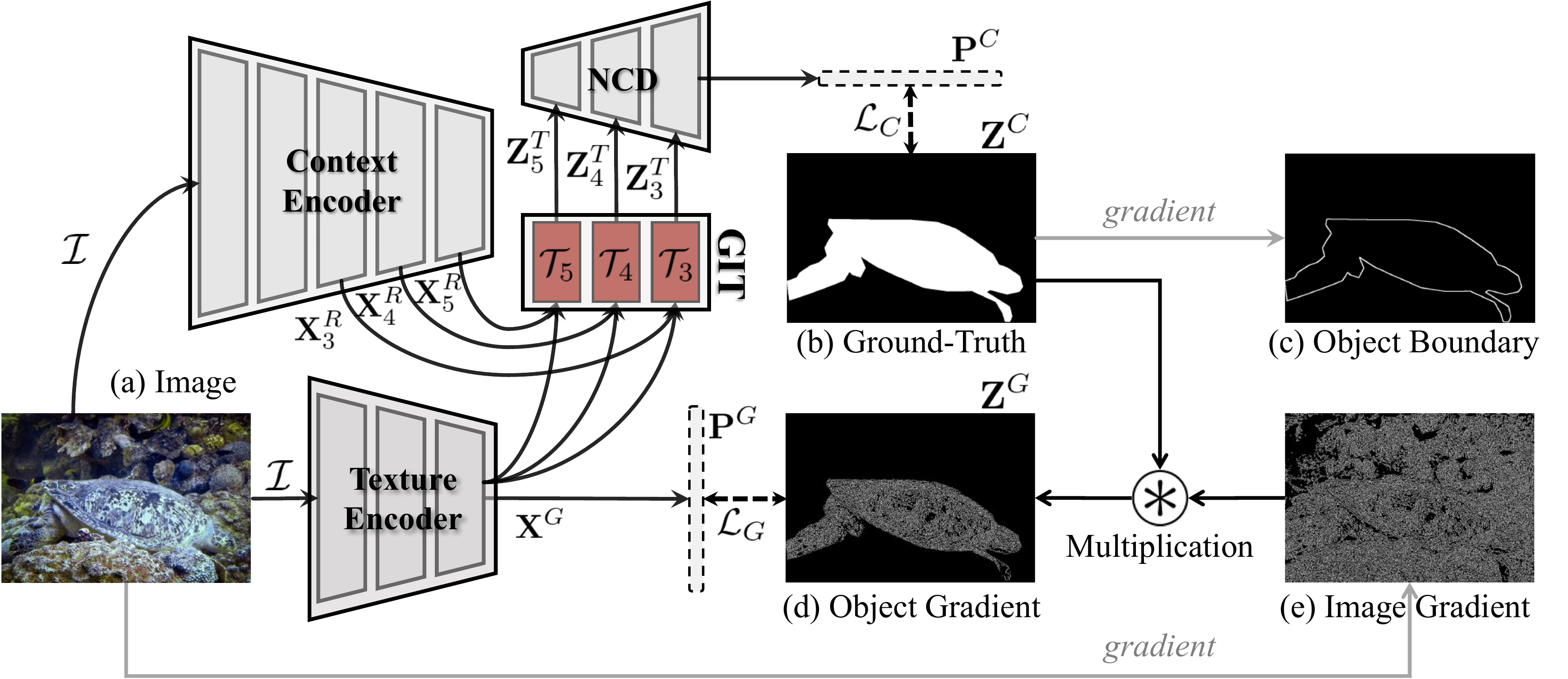}
    \end{overpic}
    \caption{Overall pipeline of the proposed \ourmodel. It consists of two connected learning branches, \ie, context encoder (\secref{sec:context_path}) and texture encoder (\secref{sec:texture_path}). Then, we introduce a gradient-induced transition (GIT) (\secref{sec:GIDecoder}) to collaboratively aggregate the feature that is derived from the above two encoders. Finally, a neighbor connected decoder (NCD)~\cite{fan2021concealed} is adopted to generate the prediction $\mathbf{P}^C$ (\secref{sec:architecture_details}).}
    \label{fig:dgnet}
\end{figure*}

As discussed in~\cite{lin2017feature}, the low-level and high-level features occupy an equal role in the scene understanding. As suggested by~\cite{ke2022modnet}, it is not encouraged to encode them simultaneously.
As shown in~\figref{fig:dgnet}, we propose to model the camouflaged representations with two separate encoders, a context and a texture encoder. 

\subsection{Context Encoder}\label{sec:context_path}

For a camouflaged input image $\mathcal{I} \in \mathbb{R}^{3 \times H \times W}$, we use the widely used EfficientNet~\cite{tan2019efficientnet} as the context encoder to obtain the \jgp{pyramid features} $\{\mathbf{X}_i\}_{i=1}^5$.

\noindent\textbf{Dimensional Reduction.}
Inspired by~\cite{fan2020pranet}, we adopt the following two steps to ensure efficient element-wise operations between different levels in the decoding stage:
\textit{a)} we only pick out the top-three features (\ie, when $i=3,4,5$), which retain the affluent semantics of the visual scene.
\textit{b)} we further utilize two stacked \texttt{ConvBR}\footnote{In this paper, \texttt{ConvBR} denotes the standard convolutional layer followed by a BN~\cite{ioffe2015batch} layer and a ReLU~\cite{glorot2011deep} layer.} layers with $C_i\times3\times3$ filters to reduce the dimension of each candidate feature to $C_i$, contributing to easing the computational burden of subsequent operations.
The final outputs are three context features $\{\mathbf{X}_i^R\}_{i=3}^5 \in \mathbb{R}^{C_i \times H_i \times W_i}$, where $C_i$, $H_i = \frac{H}{2^i}$, and $W_i = \frac{W}{2^i}$ denote the channel, height, and width of the feature maps.

\subsection{Texture Encoder}\label{sec:texture_path}
We also introduce a tailored texture branch supervised by the object-level gradient map, compensating for the pattern degradation caused by the top-three context features' weak representation of geometric textures.

\noindent\textbf{Object Gradient Generation.}
An image gradient describes the directional change in an image's intensity or color between adjacent positions, which is widely applied for edge detection~\cite{su2021pixel} and super-resolution~\cite{ma2020structure}.
The right part of~\figref{fig:dgnet} presents four widely used types of supervision labels.
The object boundary (c) and image gradient (e) can be directly generated by calculating the gradient of the object-level ground-truth $\mathbf{Z}^C$ (b) and raw image (a), respectively.
However, the raw image gradient map (e), which contains irrelevant background noises, may mislead the optimization process when serving as the supervision signal for texture learning.
To address this problem, we introduce a novel camouflage learning paradigm that uses the object-level gradient map $\mathbf{Z}^G$ (d) as supervision, which holds both the gradient cues of the object's boundaries and interior regions.
This process could be formulated as:
\begin{equation}\label{equ:grad_descriptor}
    \mathbf{Z}^G = \mathcal{F}_E (\mathcal{I}(x,y)) \circledast \mathbf{Z}^C,
\end{equation}
where $\mathcal{F}_E$ represents the standard Canny edge detector~\cite{canny1986computational} for input $\mathcal{I}$ with discrete pixel coordinates $(x,y)$.
$\circledast$ means the element-wise multiplication.

\begin{table}[t!]
    \centering
    \footnotesize
    \caption{Details of the tailored texture encoder.  
    $k$: kernel size, $c$: output channels, $s$: stride, and $p$: zero-padding. Here, we set the channel $C_g=32$ as default setting.}
    \label{tab:texture_encoder}
    \renewcommand{\arraystretch}{1.5}
    \renewcommand{\tabcolsep}{2.0pt}
    \begin{tabular}{c||r|r|c|c|c|c|c}
    \hline
    Layer &Input Size & Output Size & Component &$k$ &$c$ &$s$ &$p$  \\
    \hline
    $\#01$ &$3\times H \times W$ &$64 \times \frac{H}{2} \times \frac{W}{2}$ &\texttt{ConvBR} &7 &64 &2 &3 \\
    $\#02$ &$64 \times \frac{H}{2} \times \frac{W}{2}$ &$64 \times \frac{H}{4} \times \frac{W}{4}$ &\texttt{ConvBR} &3 &64 &2 &1  \\
    $\#03$ &$64\times \frac{H}{4} \times \frac{W}{4}$ &$C_g \times \frac{H}{8} \times \frac{W}{8}$ &\texttt{ConvBR} &3 &$C_g$ &2 &1 \\
    $\#04$ &$C_g \times \frac{H}{8} \times \frac{W}{8}$ &$1 \times \frac{H}{8} \times \frac{W}{8}$ &\texttt{ConvBR} &1 &1 &1 &0 \\
    \hline
    \end{tabular}
\end{table}

\noindent\textbf{Texture Encoder.}
Because low-level features with a high resolution will introduce a computational burden, we design a tailored lightweight encoder instead of utilizing an out-of-box backbone. 
We obtain the texture feature $\mathbf{X}^G \in \mathbb{R}^{C_g \times H_g \times W_g}$ from layer$\#03$ (see~\tabref{tab:texture_encoder}). However, we supervise the following layer$\#04$ with the object-level gradient $\mathbf{Z}^G$.
We keep the texture feature with a larger resolution (\ie, $H_g = \frac{H}{8}$ and $W_g = \frac{W}{8}$) since the features with smaller resolution would discard most geometric details.

\subsection{Gradient-Induced Transition}\label{sec:GIDecoder}
\begin{figure*}[t!]
    \centering
    \begin{overpic}[width=.95\linewidth]{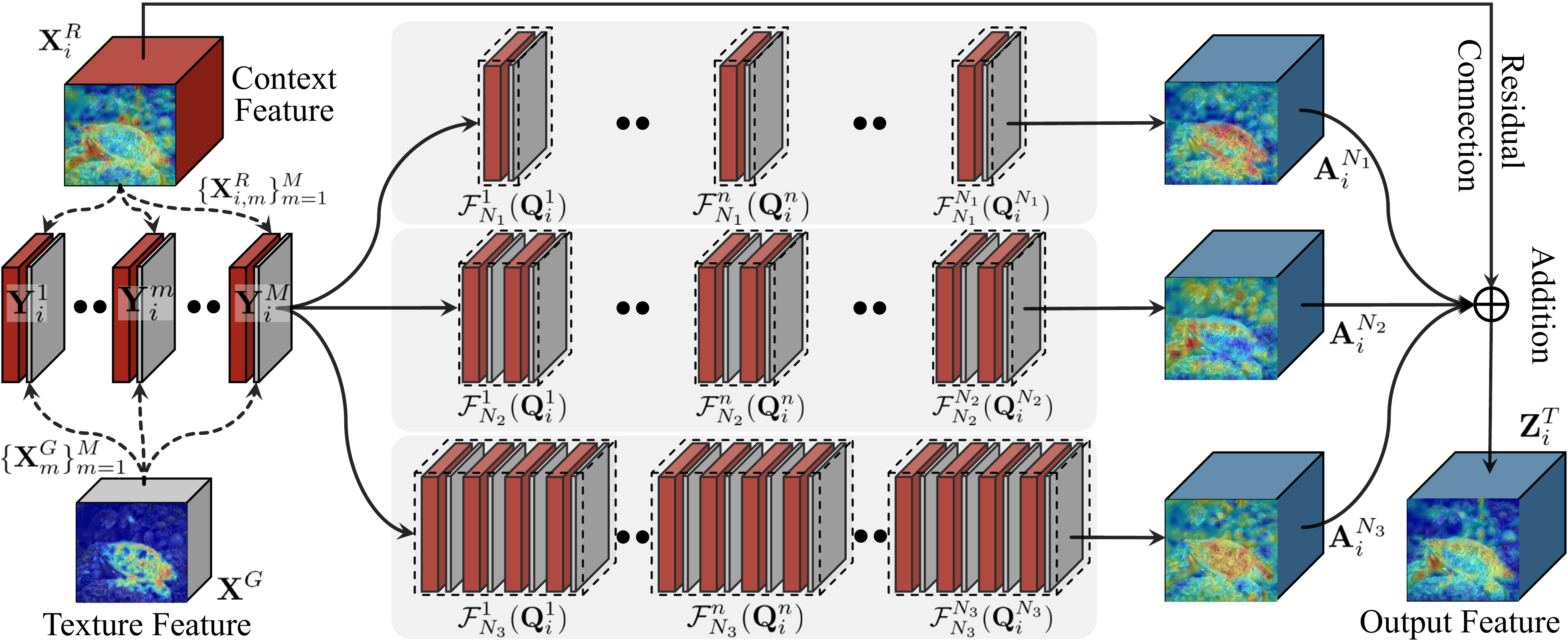}
    \end{overpic}
    \caption{Illustration of the proposed gradient-induced transition (GIT).}
    \label{fig:git}
\end{figure*}

The latent correlation between context and texture features offers great potential for adaptive fusion rather than adopting naive fusion strategies (\eg, concatenation and addition operations).
Here, we design a flexible plug-and-play \textit{gradient-induced transition (GIT)} module (see~\figref{fig:git}), which views the texture feature as the auxiliaries in the multi-source aggregation from a group-wise perspective.
Specifically, it comprises the following three steps.

\noindent\textbf{Gradient-Induced Group Learning.}
Inspired by~\cite{fan2021concealed}, we first adopt the gradient-induced group learning strategy, which splits three context features $\{\mathbf{X}^R_i\}_{i=3}^5$ and a texture feature $\mathbf{X}^G$ into fixed groups along the channel dimension.
This strategy can be formulated for each $\mathbf{X}^R_i$ and $\mathbf{X}^G$ pair:
\begin{equation}\label{equ:group_number}
\begin{split}
    \{ \mathbf{X}_{i,m}^R \}_{m=1}^M \in \mathbb{R}^{K_i \times H_i \times W_i} \leftarrow \mathbf{X}^R_i \in \mathbb{R}^{C_i \times H_i \times W_i},\\
    \{ \mathbf{X}_m^G \}_{m=1}^M \in \mathbb{R}^{K_g \times H_g \times W_g} \leftarrow \mathbf{X}^G \in \mathbb{R}^{C_g \times H_g \times W_g},
\end{split}
\end{equation}
where $\leftarrow$ is the feature grouping operation.
$K_i=C_i/M$ and $K_g=C_g/M$ denote the channel number of each feature group, and $M$ is the corresponding number of groups.
Then, we periodically arrange groups of context features $\mathbf{X}_{i,m}^R$ and texture features $\mathbf{X}^G_m$, which generates the regrouped feature $\mathbf{Q}_i$ \jgp{via}:
\begin{equation}\label{equ:regroup_feature}
    \mathbf{Q}_i \in \mathbb{R}^{(C_i + C_g) \times H_i \times W_i} = \langle \mathbf{Y}^1_i ; \dots ; \mathbf{Y}^m_i ; \dots ; \mathbf{Y}^M_i \rangle,
\end{equation}
where $\langle \cdot;\cdot \rangle$ means the channel-wise feature concatenation.
Here, the m-\textit{th} sub-component $\mathbf{Y}^m_i$ is derived from:
\begin{equation}
\mathbf{Y}^m_i \in \mathbb{R}^{(K_i+K_g) \times H_i \times W_i} = \langle  \mathcal{F}_{\downarrow}({\mathbf{X}^G_m}); \mathbf{X}_{i,m}^R \rangle,
\end{equation}
where the down-sampling operation $\mathcal{F}_{\downarrow}(\cdot)$ ensures the spatial resolution of $\mathbf{X}_m^G$ matches $\mathbf{X}_{i,m}^R$.

\noindent\textbf{Soft Grouping Strategy.}
The naive feature fusion strategies may ignore the correlation or distinctiveness between context and texture representations due to lacking further multi-source interactions.
Inspired by the parallel design introduced in~\cite{chen2017deeplab} \jgp{for capturing} objects at multiple scales, we propose using a soft grouping strategy to provide parallel nonlinear projections at multiple fine-grained sub-spaces, which enables the network to probe multi-source representations jointly.
\res{Specifically, we set three parallel sub-branches (\ie, $\{N_1,N_2,N_3\}$ as in the gray region of \figref{fig:git}) for the soft grouping in our experiment. Here, for simplifying illustration, we take the $N$-th sub-branch as an example via omitting the subscript, which could be formulated as:}
\begin{equation}\label{equ:group_att}
    \mathbf{A}_i^N = \langle \mathcal{F}^1_{N}(\mathbf{Q}_i^1); \dots; \mathcal{F}^n_{N}(\mathbf{Q}_i^n) \dots \mathcal{F}^N_{N}(\mathbf{Q}_i^N) \rangle,
\end{equation}
where $\mathcal{F}^n_{N}(\mathbf{Q}_i^n) \in \mathbb{R}^{(C_i/N) \times H_i \times W_i} = f_n(\mathbf{Q}_i^n,\omega_n)$ intentionally introduces soft non-linearity at each  multi-source sub-space.
The projection function $f_n$ is implemented by a convolutional layer \jgp{with $C_i$ filters of shape of $\frac{(C_i+C_g)}{N}\times1\times1$}, which is parameterized by learnable weights $\omega_n$.
Here, $\mathbf{Q}_i^n$ is the n-\textit{th} subset of the regrouped feature $\mathbf{Q}_i$ that is divided into $N$ groups.

\noindent\textbf{Parallel Residual Learning.}
We further introduce residual learning~\cite{he2016deep} in a parallel manner at different group-aware scales.
Consequently, we can define the GIT function $\mathcal{T}_i(\cdot,\cdot)$ (see the red block in~\figref{fig:dgnet}) as:
\begin{equation}\label{equ:transition}
    \mathbf{Z}_i^T = \mathcal{T}_i(\mathbf{X}^R_i,\mathbf{X}^G) = \mathbf{X}_i^R \oplus \sum_{N} \mathbf{A}_i^{N},
\end{equation}
where $N \in \{N_1, N_2, N_3\}$ denotes a set of scaling factors for different groups, which will be discussed in~\secref{sec:ablation}.
$\oplus$ means the element-wise addition, and $\sum$ denotes a sum of multiple terms.
The final output is $\{\mathbf{Z}_i^T\}_{i=3}^5 \in \mathbb{R}^{C_i \times H_i \times W_i}$.

\subsection{Learning Details}\label{sec:architecture_details}

\noindent\textbf{Decoder.}
Given the context features $\{\mathbf{X}_i^R\}_{i=3}^{5}$, we firstly apply the GIT function $\mathcal{T}_i(\cdot,\cdot)$ (see~\equref{equ:transition}) to get the output features $\{ \mathbf{Z}_i^T\}_{i=3}^{5}$.
To exploit the above gradient-induced features in $\mathbf{Z}^T_i$ more efficiently, we utilize the neighbour connection decoder (NCD)~\cite{fan2021concealed} to generate the final prediction, enabling feature propagation from high to low levels.
Thus, the final prediction $\mathbf{P}^C$ can be derived from $\mathbf{P}^C \in \mathbb{R}^{1 \times H \times W} =\text{NCD}(\mathbf{Z}_3^T, \mathbf{Z}_4^T, \mathbf{Z}_5^T)$.

\noindent\textbf{Loss Function.}
The overall optimization objective is defined as:
\begin{equation}
    \mathcal{L}=\mathcal{L}_{C}(\mathbf{P}^C,\mathbf{Z}^C)+\mathcal{L}_{G}(\mathbf{P}^G,\mathbf{Z}^G),
\end{equation}
where $\mathcal{L}_{C}$ and $\mathcal{L}_{G}$ represent the \res{segmentation and object gradient loss functions, respectively.}
For the former, it is formulated as $\mathcal{L}_{C}=\mathcal{L}_{IoU}^{w}+\mathcal{L}_{BCE}^{w}$, where $\mathcal{L}_{IoU}^{w}$ and $\mathcal{L}_{BCE}^{w}$ represent the weighted intersection-over-union (IoU) loss and the weighted binary cross-entropy (BCE) loss, respectively.
They assign the adaptive weight for each pixel according to its difficulty in focusing on the global structure and paying more attention to the hard pixels. 
The definitions of these losses are the same as in~\cite{wei2020f3net,fan2020camouflaged,fan2021concealed} and their effectiveness has been proven in binary segmentation.
For the latter, we employ the standard mean squared error loss function.

\begin{table}[t!]
    \centering
    \footnotesize
    \caption{Hyper-parameter settings of the proposed \ourmodelS~and~\ourmodel.}
    \label{tab:hyper_parameter}
    \renewcommand{\arraystretch}{1.7}
    \renewcommand{\tabcolsep}{4.0pt}
    \begin{tabular}{l||c|c|c|c|c}
    \hline
    Model &Backbone &$C_i$ &$C_g$ &$M$ &$\{N_1,N_2,N_3\}$ \\
    \hline
    \textbf{\ourmodelS} &EfficientNet-B1 &32 &32 &8 &$\{8,16,32\}$ \\
    \textbf{\ourmodelL} &EfficientNet-B4 &64 &32 &8 &$\{4,8,16\}$ \\
    \hline
    \end{tabular}
\end{table}

\noindent\textbf{Training Settings.}
The proposed \ourmodel~is implemented in the \jgp{PyTorch~\cite{paszke2019pytorch}/Jittor~\cite{hu2020jittor}} toolbox and trained/inferred on a single NVIDIA RTX TITAN GPU.
The model parameters are initialized with the strategy of~\cite{he2015delving}, and we initialize the backbone with the model weights pre-trained on ImageNet~\cite{krizhevsky2012imagenet} to prevent over-fitting.
We discard the last stage of Conv1$\times$1, pooling, and fully connected layers in the EfficientNet~\cite{tan2019efficientnet} backbone and extract the features from the top-three lateral outputs, including \texttt{stage-4} ($\mathbf{X}_3$), \texttt{stage-6} ($\mathbf{X}_4$), and \texttt{stage-8} ($\mathbf{X}_5$).
Considering the performance-efficiency trade-off, we instantiate two variants to adapt the specific requirement under various computational overheads (refer to~\tabref{tab:hyper_parameter}).

We train our model in an end-to-end manner, using Adam~\cite{kingma2015adam}.
The cosine annealing part of the SGDR strategy~\cite{loshchilov2017sgdr} is used to adjust the learning rate, where the minimum/maximum learning rate and the maximum adjusted iteration are set to $10^{-5}$/$10^{-4}$ and 20, respectively.
The batch size is set to 12, and the maximum training epoch is 100.
During training, we resize each image to 352$\times$352 and feed it into \ourmodel~with four data augmentation techniques: color enhancement, random flipping, random cropping, and random rotation.
\res{Finally, our DGNet and DGNet-S take 8.8 and 7.9 hours to reach the network convergence.}

\noindent\textbf{Testing Settings.} 
Once the network is well-trained, we resize the input images to 352$\times$352 and test our \ourmodelS~and \ourmodelL~on three unseen test datasets. 
We take the final output $\mathbf{P}^C$ as the prediction map without any heuristic post-processing techniques, such as DenseCRF~\cite{krahenbuhl2011efficient}.

\section{Experiments}

\subsection{Benchmarking}

\noindent\textbf{Datasets.} 
There are three popular datasets in the COD field:
\textit{a)} CAMO~\cite{le2019anabranch} has 1,250 camouflaged images and is divided into CAMO-\texttt{Tr} (1,000 samples) and CAMO-\texttt{Te} (250 samples).
\textit{b)} COD10K~\cite{fan2021concealed} is the largest COD dataset till now, consisting of COD10K-\texttt{Tr} (3,040 images) and COD10K-\texttt{Te} (2,026 images). It is downloaded from multiple free photography websites, covering 5 super-classes and 69 sub-classes.
\textit{c)} NC4K-\texttt{Te}~\cite{lyu2021simultaneously}, as the largest testing dataset, includes 4,121 samples, which \jgp{are used to evaluate the models' generalization ability}.
Following the protocol of~\cite{fan2021concealed}, we train our model on the hybrid dataset (\ie, COD10K-\texttt{Tr} + CAMO-\texttt{Tr}) with 4,040 samples and evaluate our method on above three benchmarks (see~\tabref{tab:fullmetric_rank}).

\noindent\textbf{Metrics.}
Following~\cite{fan2021concealed}, we use five commonly used metrics for the evaluation: structure measure ($\mathcal{S}_{\alpha}$)~\cite{fan2017structure}, \jgp{enhanced-alignment measure ($E_{\phi}$)~\cite{Fan2018Enhanced,fan2021cognitive}, F-measure ($F_{\beta}$})~\cite{borji2015salient,zhuge2021salient}, weighted F-measure ($F_{\beta}^{w}$)~\cite{margolin2014evaluate}, and mean absolute error ($\mathcal{M}$).
\jgp{Besides, the precision-recall (PR) curves~\cite{borji2015salient} are obtained by varying different thresholds from $[0,255]$. Similar to this thresholding strategy, F-measure and E-measure curves are also reported.
Moreover, we adopt three criteria} to measure the model's complexity\footnote{The model's parameter and MACs are measured by the toolbox: \url{https://github.com/sovrasov/flops-counter.pytorch}.} and efficiency:
the number of model parameters, measured in Millions (M), 
the number of multiply-accumulate (MACs) operations, measured in Giga (G),
and inference speed measured in frames per second (fps).

\begin{table*}[ht!]
    \centering
    \footnotesize
    \caption{\jgp{Quantitative results in terms of full metrics for cutting-edge competitors, including 8 SOD-related and 12 COD-related, on three COD-related test datasets. @$\mathcal{R}$ is the ranking of the current metric, and Mean@$\mathcal{R}$ indicates the mean ranking of all metrics. Note that $E_\phi^{mx}$/$F_\beta^{mx}$, $E_\phi^{mn}$/$F_\beta^{mn}$, and $E_\phi^{ad}$/$F_\beta^{ad}$ denote the maximum, mean, and adaptive value of E-measure/F-measure, respectively. $\uparrow$/$\downarrow$ denotes that the higher/lower the score, the better. All the benchmark results are available at \href{OneDrive-3.54 GB}{https://anu365-my.sharepoint.com/:u:/g/personal/u7248002_anu_edu_au/EXLiBgp9nGNApBw9im2xznsBJ_ryGEW7hkJlL92gNaRAAg?e=XIUBnq}.}}
    \label{tab:fullmetric_rank}
    \renewcommand\arraystretch{1.05} 
    \setlength\tabcolsep{2.05pt} 
    \begin{tabular}{r|r||cccccccc|cccccccccccc|cc} 
    \hline 
    & & \multicolumn{8}{c|}{\tabincell{c}{SOD-related Models}} & \multicolumn{12}{c|}{\tabincell{c}{COD-related Models}} & \\
    \hline
    & 
    &\rotatebox{90}{EGNet} &\rotatebox{90}{SCRN} &\rotatebox{90}{CPD}  &\rotatebox{90}{CSNet-R} &\rotatebox{90}{F3Net} &\rotatebox{90}{UCNet} &\rotatebox{90}{ITSD} &\rotatebox{90}{MINet} 
    &\rotatebox{90}{SINet} &\rotatebox{90}{PraNet} &\rotatebox{90}{BAS} &\rotatebox{90}{C2FNet} &\rotatebox{90}{TINet} &\rotatebox{90}{UGTR} &\rotatebox{90}{PFNet} &\rotatebox{90}{S-MGL} &\rotatebox{90}{R-MGL} &\rotatebox{90}{LSR} &\rotatebox{90}{JCSOD} &\rotatebox{90}{SINetV2} &\rotatebox{90}{\textbf{\ourmodelS}} &\rotatebox{90}{\textbf{\ourmodel}}
    \\
    & Metric
    &\cite{zhao2019EGNet} &\cite{wu2019stacked} &\cite{wu2019cascaded} &\cite{gao2020highly} &\cite{wei2020f3net} &\cite{zhang2020UCNet} &\cite{zhou2020interactive} &\cite{pang2020multi} 
    &\cite{fan2020camouflaged} &\cite{fan2020pranet} &\cite{qin2021boundary} &\cite{sun2021Context} &\cite{zhu2021inferring} &\cite{yang2021uncertainty} &\cite{mei2021camouflaged} &\cite{zhai2021mutual} &\cite{zhai2021mutual} &\cite{lyu2021simultaneously} &\cite{li2021uncertainty} &\cite{fan2021concealed}
    &\multicolumn{2}{c}{\tabincell{c}{\textbf{\textit{Ours}}}}
    \\
    \hline
    \multirow{18}{*}{\begin{sideways}NC4K-\texttt{Te}~\cite{lyu2021simultaneously}\end{sideways}}
    & $\mathcal{S}_{\alpha}\uparrow$
    &.777 &.830 &.788 &.750 &.780 &.811 &.811 &.805 &.808 &.822 &.817 &.838 &.829 &.839 &.829 &.829 &.833 &.840 &.842 &\cellcolor{myGreen!30}.847 &\cellcolor{myBlue!30}.845 &\cellcolor{myRed!30}.857 
    \\
    & @$ \mathcal{R}$
    &21 &9 &19 &22 &20 &16 &15 &18 &17 &13 &14 &7 &12 &6 &11 &10 &8 &5 &4 &\tg{2} &\tb{3} &\tr{1} 
    \\
    & $E_\phi^{mx}\uparrow$
    &.864 &.897 &.865 &.793 &.848 &.886 &.883 &.881 &.883 &.888 &.872 &.904 &.890 &.899 &.898 &.893 &.893 &.907 &.907 &\cellcolor{myGreen!30}.914 &\cellcolor{myBlue!30}.913 &\cellcolor{myRed!30}.922 
    \\
    & @$ \mathcal{R}$
    &20 &9 &19 &22 &21 &14 &15 &17 &16 &13 &18 &6 &12 &7 &8 &11 &10 &5 &4 &\tg{2} &\tb{3} &\tr{1} 
    \\
    & $E_\phi^{mn}\uparrow$
    &.841 &.854 &.804 &.773 &.824 &.871 &.845 &.846 &.871 &.876 &.859 &.897 &.879 &.874 &.888 &.863 &.867 &.895 &.898 &\cellcolor{myGreen!30}.903 &\cellcolor{myBlue!30}.902 &\cellcolor{myRed!30}.911 
    \\
    & @$ \mathcal{R}$
    &19 &16 &21 &22 &20 &11 &18 &17 &12 &9 &15 &5 &8 &10 &7 &14 &13 &6 &4 &\tg{2} &\tb{3} &\tr{1} 
    \\
    & $E_\phi^{ad}\uparrow$
    &.826 &.864 &.842 &.812 &.853 &.883 &.855 &.876 &.882 &.871 &.868 &.898 &.880 &.886 &.892 &.884 &.889 &\cellcolor{myBlue!30}.902 &\cellcolor{myGreen!30}.903 &.898 &.899 &\cellcolor{myRed!30}.907 
    \\
    & @$ \mathcal{R}$
    &21 &17 &20 &22 &19 &11 &18 &14 &12 &15 &16 &6 &13 &9 &7 &10 &8 &\tb{3} &\tg{2} &5 &4 &\tr{1} 
    \\
    & $F_\beta^w\uparrow$
    &.639 &.698 &.632 &.603 &.656 &.729 &.680 &.705 &.723 &.724 &.732 &.762 &.734 &.747 &.745 &.731 &.740 &.766 &\cellcolor{myGreen!30}.771 &\cellcolor{myBlue!30}.770 &.764 &\cellcolor{myRed!30}.784 
    \\
    & @$ \mathcal{R}$
    &20 &17 &21 &22 &19 &13 &18 &16 &15 &14 &11 &6 &10 &7 &8 &12 &9 &4 &\tg{2} &\tb{3} &5 &\tr{1} 
    \\
    & $F_\beta^{mx}\uparrow$
    &.731 &.793 &.738 &.669 &.719 &.782 &.762 &.768 &.775 &.786 &.782 &.810 &.793 &.807 &.799 &.797 &.800 &.815 &.816 &\cellcolor{myGreen!30}.823 &\cellcolor{myBlue!30}.819 &\cellcolor{myRed!30}.833 
    \\
    & @$ \mathcal{R}$
    &20 &11 &19 &22 &21 &15 &18 &17 &16 &13 &14 &6 &12 &7 &9 &10 &8 &5 &4 &\tg{2} &\tb{3} &\tr{1} 
    \\
    & $F_\beta^{mn}\uparrow$
    &.696 &.757 &.695 &.655 &.705 &.775 &.729 &.753 &.769 &.762 &.772 &.795 &.773 &.787 &.784 &.777 &.782 &.804 &\cellcolor{myGreen!30}.806 &\cellcolor{myBlue!30}.805 &.799 &\cellcolor{myRed!30}.814 
    \\
    & @$ \mathcal{R}$
    &20 &16 &21 &22 &19 &11 &18 &17 &14 &15 &13 &6 &12 &7 &8 &10 &9 &4 &\tg{2} &\tb{3} &5 &\tr{1} 
    \\
    & $F_\beta^{ad}\uparrow$
    &.671 &.744 &.709 &.672 &.710 &.776 &.717 &.763 &.768 &.753 &.767 &.788 &.766 &.779 &.779 &.771 &.778 &\cellcolor{myBlue!30}.802 &\cellcolor{myGreen!30}.803 &.792 &.789 &\cellcolor{myRed!30}.803 
    \\
    & @$ \mathcal{R}$
    &22 &17 &20 &21 &19 &10 &18 &15 &12 &16 &13 &6 &14 &8 &7 &11 &9 &\tb{3} &\tg{2} &4 &5 &\tr{1} 
    \\
    & $\mathcal{M}\downarrow$
    &.075 &.059 &.074 &.088 &.070 &.055 &.064 &.060 &.058 &.059 &.058 &.049 &.055 &.052 &.053 &.055 &.052 &.048 &\cellcolor{myGreen!30}.047 &.048 &\cellcolor{myBlue!30}.047 &\cellcolor{myRed!30}.042 
    \\
    & @$ \mathcal{R}$
    &21 &15 &20 &22 &19 &10 &18 &17 &14 &16 &13 &6 &12 &8 &9 &11 &7 &5 &\tg{2} &4 &\tb{3} &\tr{1} 
    \\
    \hline
    & Mean@$ \mathcal{R}$
    &21 &15 &20 &22 &19 &12 &18 &17 &16 &13 &14 &6 &11 &7 &8 &10 &9 &5 &\tg{2} &\tb{3} &4 &\tr{1} 
    \\
    \hline
    \multirow{18}{*}{\begin{sideways}CAMO-\texttt{Te}~\cite{le2019anabranch}\end{sideways}}
    & $\mathcal{S}_{\alpha}\uparrow$
    &.732 &.779 &.726 &.771 &.711 &.739 &.750 &.737 &.745 &.769 &.749 &.796 &.781 &.785 &.782 &.772 &.775 &.787 &.800 &\cellcolor{myBlue!30}.820 &\cellcolor{myGreen!30}.826 &\cellcolor{myRed!30}.839 
    \\
    & @$ \mathcal{R}$
    &20 &10 &21 &13 &22 &18 &15 &19 &17 &14 &16 &5 &9 &7 &8 &12 &11 &6 &4 &\tb{3} &\tg{2} &\tr{1} 
    \\
    & $E_\phi^{mx}\uparrow$
    &.820 &.850 &.801 &.849 &.780 &.820 &.830 &.818 &.829 &.837 &.808 &.864 &.848 &.854 &.855 &.842 &.842 &.854 &.873 &\cellcolor{myBlue!30}.895 &\cellcolor{myGreen!30}.907 &\cellcolor{myRed!30}.915 
    \\
    & @$ \mathcal{R}$
    &17 &9 &21 &10 &22 &18 &15 &19 &16 &14 &20 &5 &11 &8 &6 &13 &12 &7 &4 &\tb{3} &\tg{2} &\tr{1} 
    \\
    & $E_\phi^{mn}\uparrow$
    &.800 &.797 &.723 &.795 &.741 &.787 &.780 &.767 &.804 &.824 &.796 &.854 &.836 &.823 &.842 &.807 &.812 &.838 &.859 &\cellcolor{myBlue!30}.882 &\cellcolor{myGreen!30}.893 &\cellcolor{myRed!30}.901 
    \\
    & @$ \mathcal{R}$
    &14 &15 &22 &17 &21 &18 &19 &20 &13 &9 &16 &5 &8 &10 &6 &12 &11 &7 &4 &\tb{3} &\tg{2} &\tr{1} 
    \\
    & $E_\phi^{ad}\uparrow$
    &.811 &.848 &.810 &.847 &.802 &.811 &.830 &.826 &.825 &.833 &.806 &.864 &.845 &.859 &.852 &.850 &.847 &.855 &.865 &\cellcolor{myBlue!30}.875 &\cellcolor{myGreen!30}.892 &\cellcolor{myRed!30}.901 
    \\
    & @$ \mathcal{R}$
    &18 &10 &20 &12 &22 &19 &15 &16 &17 &14 &21 &5 &13 &6 &8 &9 &11 &7 &4 &\tb{3} &\tg{2} &\tr{1} 
    \\
    & $F_\beta^w\uparrow$
    &.604 &.643 &.553 &.642 &.564 &.640 &.610 &.613 &.644 &.663 &.646 &.719 &.678 &.686 &.695 &.664 &.673 &.696 &.728 &\cellcolor{myBlue!30}.743 &\cellcolor{myGreen!30}.754 &\cellcolor{myRed!30}.769 
    \\
    & @$ \mathcal{R}$
    &20 &15 &22 &16 &21 &17 &19 &18 &14 &12 &13 &5 &9 &8 &7 &11 &10 &6 &4 &\tb{3} &\tg{2} &\tr{1} 
    \\
    & $F_\beta^{mx}\uparrow$
    &.688 &.738 &.667 &.740 &.630 &.708 &.694 &.683 &.708 &.728 &.703 &.771 &.745 &.754 &.758 &.739 &.740 &.753 &.779 &\cellcolor{myBlue!30}.801 &\cellcolor{myGreen!30}.810 &\cellcolor{myRed!30}.822 
    \\
    & @$ \mathcal{R}$
    &19 &13 &21 &10 &22 &15 &18 &20 &16 &14 &17 &5 &9 &7 &6 &12 &11 &8 &4 &\tb{3} &\tg{2} &\tr{1} 
    \\
    & $F_\beta^{mn}\uparrow$
    &.670 &.705 &.614 &.705 &.616 &.700 &.663 &.667 &.702 &.710 &.692 &.762 &.728 &.738 &.746 &.721 &.726 &.744 &.772 &\cellcolor{myBlue!30}.782 &\cellcolor{myGreen!30}.792 &\cellcolor{myRed!30}.806 
    \\
    & @$ \mathcal{R}$
    &18 &13 &22 &14 &21 &16 &20 &19 &15 &12 &17 &5 &9 &8 &6 &11 &10 &7 &4 &\tb{3} &\tg{2} &\tr{1} 
    \\
    & $F_\beta^{ad}\uparrow$
    &.667 &.733 &.678 &.730 &.661 &.716 &.692 &.704 &.712 &.715 &.696 &.764 &.729 &.749 &.751 &.733 &.738 &.756 &.779 &\cellcolor{myBlue!30}.779 &\cellcolor{myGreen!30}.786 &\cellcolor{myRed!30}.804 
    \\
    & @$ \mathcal{R}$
    &21 &10 &20 &12 &22 &14 &19 &17 &16 &15 &18 &5 &13 &8 &7 &11 &9 &6 &4 &\tb{3} &\tg{2} &\tr{1} 
    \\
    & $\mathcal{M}\downarrow$
    &.109 &.090 &.114 &.092 &.109 &.094 &.102 &.096 &.092 &.094 &.096 &.080 &.087 &.086 &.085 &.089 &.088 &.080 &.073 &\cellcolor{myBlue!30}.070 &\cellcolor{myGreen!30}.063 &\cellcolor{myRed!30}.057 
    \\
    & @$ \mathcal{R}$
    &21 &12 &22 &13 &20 &15 &19 &18 &14 &16 &17 &6 &9 &8 &7 &11 &10 &5 &4 &\tb{3} &\tg{2} &\tr{1} 
    \\
    \hline
    & Mean@$ \mathcal{R}$
    &20 &12 &21 &13 &22 &16 &18 &19 &15 &14 &17 &5 &9 &8 &7 &11 &10 &6 &4 & \tb{3} & \tg{2} & \tr{1} 
    \\
    \hline
    \multirow{18}{*}{\begin{sideways}COD10K-\texttt{Te}~\cite{fan2021concealed}\end{sideways}}
    & $\mathcal{S}_{\alpha}\uparrow$
    &.736 &.789 &.748 &.778 &.739 &.776 &.767 &.769 &.776 &.789 &.802 &.813 &.793 &\cellcolor{myGreen!30}.818 &.800 &.811 &.814 &.804 &.809 &\cellcolor{myBlue!30}.815 &.810 &\cellcolor{myRed!30}.822 
    \\
    & @$ \mathcal{R}$
    &22 &13 &20 &15 &21 &17 &19 &18 &16 &14 &10 &5 &12 &\tg{2} &11 &6 &4 &9 &8 &\tb{3} &7 &\tr{1} 
    \\
    & $E_\phi^{mx}\uparrow$
    &.855 &.880 &.842 &.871 &.819 &.867 &.861 &.864 &.874 &.879 &.870 &.900 &.878 &.891 &.890 &.890 &.890 &.892 &.891 &\cellcolor{myGreen!30}.906 &\cellcolor{myBlue!30}.905 &\cellcolor{myRed!30}.911 
    \\
    & @$ \mathcal{R}$
    &20 &11 &21 &15 &22 &17 &19 &18 &14 &12 &16 &4 &13 &7 &10 &9 &8 &5 &6 &\tg{2} &\tb{3} &\tr{1} 
    \\
    & $E_\phi^{mn}\uparrow$
    &.810 &.817 &.766 &.810 &.795 &.857 &.808 &.823 &.864 &.861 &.855 &\cellcolor{myGreen!30}.890 &.861 &.853 &.877 &.845 &.852 &.880 &.884 &.887 &\cellcolor{myBlue!30}.888 &\cellcolor{myRed!30}.896 
    \\
    & @$ \mathcal{R}$
    &19 &17 &22 &18 &21 &11 &20 &16 &8 &10 &12 &\tg{2} &9 &13 &7 &15 &14 &6 &5 &4 &\tb{3} &\tr{1} 
    \\
    & $E_\phi^{ad}\uparrow$
    &.753 &.789 &.768 &.791 &.818 &.867 &.787 &.837 &.867 &.839 &.869 &\cellcolor{myRed!30}.886 &.848 &.850 &.868 &.851 &.865 &\cellcolor{myBlue!30}.882 &\cellcolor{myGreen!30}.882 &.863 &.868 &.877 
    \\
    & @$ \mathcal{R}$
    &22 &19 &21 &18 &17 &8 &20 &16 &9 &15 &5 &\tr{1} &14 &13 &7 &12 &10 &\tb{3} &\tg{2} &11 &6 &4 
    \\
    & $F_\beta^w\uparrow$
    &.517 &.575 &.509 &.569 &.544 &.633 &.557 &.601 &.631 &.629 &.677 &\cellcolor{myGreen!30}.686 &.635 &.667 &.660 &.655 &.666 &.673 &\cellcolor{myBlue!30}.684 &.680 &.672 &\cellcolor{myRed!30}.693 
    \\
    & @$ \mathcal{R}$
    &21 &17 &22 &18 &20 &13 &19 &16 &14 &15 &5 &\tg{2} &12 &8 &10 &11 &9 &6 &\tb{3} &4 &7 &\tr{1} 
    \\
    & $F_\beta^{mx}\uparrow$
    &.633 &.699 &.634 &.679 &.609 &.691 &.658 &.672 &.691 &.704 &.729 &\cellcolor{myBlue!30}.743 &.712 &.742 &.725 &.733 &.738 &.732 &.738 &\cellcolor{myGreen!30}.752 &.743 &\cellcolor{myRed!30}.759 
    \\
    & @$ \mathcal{R}$
    &21 &14 &20 &17 &22 &16 &19 &18 &15 &13 &10 &\tb{3} &12 &5 &11 &8 &7 &9 &6 &\tg{2} &4 &\tr{1} 
    \\
    & $F_\beta^{mn}\uparrow$
    &.582 &.651 &.582 &.635 &.593 &.681 &.615 &.654 &.679 &.671 &.715 &\cellcolor{myGreen!30}.723 &.679 &.712 &.701 &.702 &.711 &.715 &\cellcolor{myBlue!30}.721 &.718 &.710 &\cellcolor{myRed!30}.728 
    \\
    & @$ \mathcal{R}$
    &22 &17 &21 &18 &20 &12 &19 &16 &13 &15 &6 &\tg{2} &14 &7 &11 &10 &8 &5 &\tb{3} &4 &9 &\tr{1} 
    \\
    & $F_\beta^{ad}\uparrow$
    &.526 &.593 &.555 &.589 &.588 &.673 &.573 &.639 &.667 &.640 &\cellcolor{myRed!30}.707 &\cellcolor{myBlue!30}.703 &.652 &.671 &.676 &.667 &.681 &.699 &\cellcolor{myGreen!30}.705 &.682 &.680 &.698 
    \\
    & @$ \mathcal{R}$
    &22 &17 &21 &18 &19 &10 &20 &16 &12 &15 &\tr{1} &\tb{3} &14 &11 &9 &13 &7 &4 &\tg{2} &6 &8 &5 
    \\
    & $\mathcal{M}\downarrow$
    &.061 &.047 &.058 &.047 &.051 &.042 &.051 &.043 &.043 &.045 &.038 &.036 &.042 &.035 &.040 &.037 &\cellcolor{myBlue!30}.035 &.037 &\cellcolor{myGreen!30}.035 &.037 &.036 &\cellcolor{myRed!30}.033 
    \\
    & @$ \mathcal{R}$
    &22 &18 &21 &17 &20 &13 &19 &15 &14 &16 &10 &5 &12 &4 &11 &9 &\tb{3} &8 &\tg{2} &7 &6 &\tr{1} 
    \\
    \hline
    & Mean@$ \mathcal{R}$
    &22 &16 &21 &18 &20 &14 &19 &17 &13 &15 &9 & \tg{2} &12 &8 &10 &11 &7 &6 & \tb{3} &4 &5 & \tr{1} 
    \\
    \hline
    \end{tabular}
    \vspace{-5pt}
\end{table*}

\noindent\textbf{Competitors.}
We compare our model with 20 SOTA competitors (see~\tabref{tab:fullmetric_rank}), including 8 SOD-based and 12 COD-based.
For a fair comparison, all results were either taken from the public website or produced by retraining the models on the same training dataset with default settings.

\begin{figure*}[t!]
    \centering
    \begin{overpic}[width=0.93\linewidth]{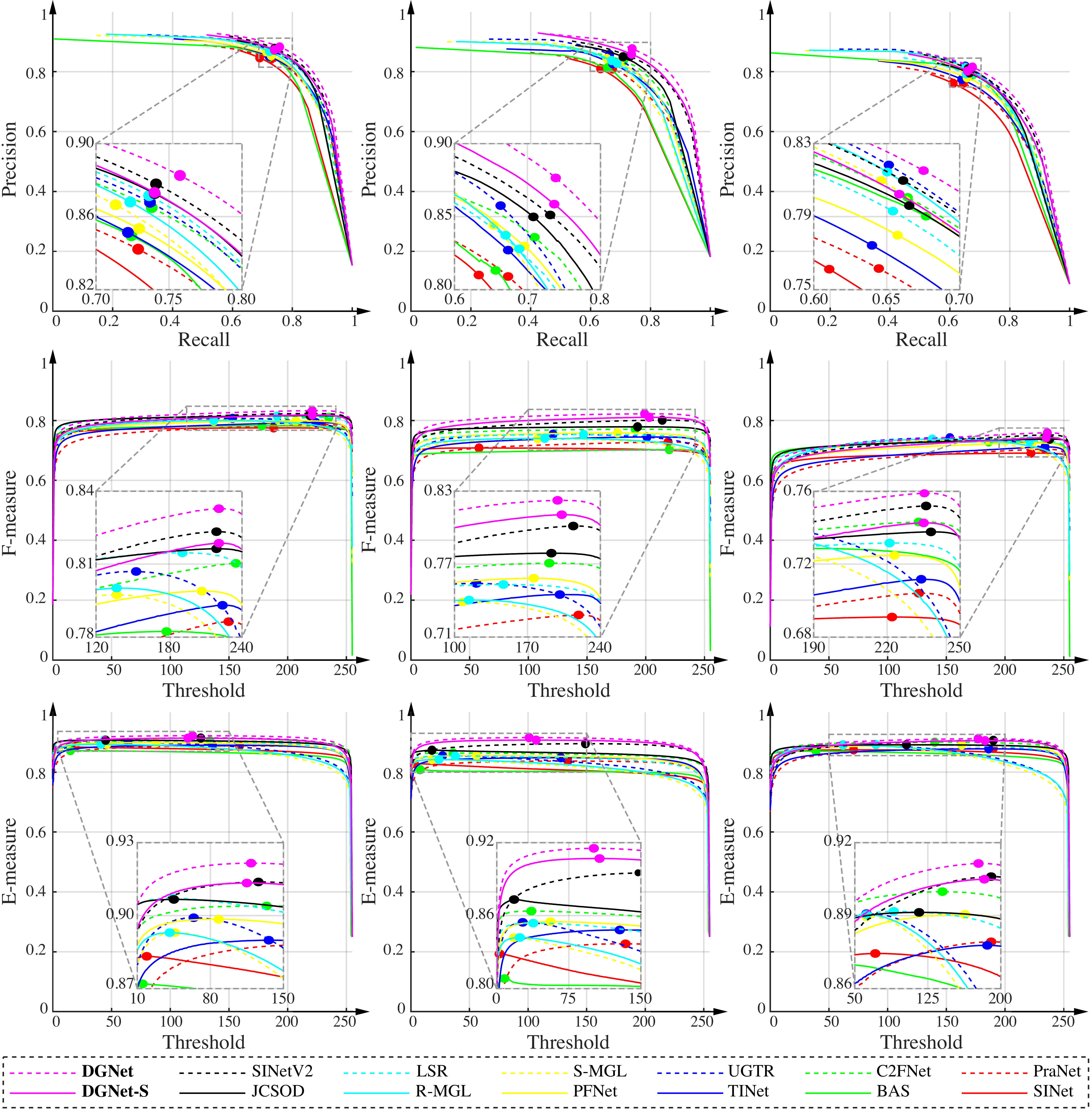}
    \put(20, 35.5){\tiny \textbf{\underline{NC4K-\texttt{Te}~\cite{lyu2021simultaneously}}}}
    \put(50.5, 35.5){\tiny \textbf{\underline{CAMO-\texttt{Te}~\cite{le2019anabranch}}}}
    \put(81, 35.5){\tiny \textbf{\underline{COD10K-\texttt{Te}~\cite{fan2021concealed}}}}
    \put(20, 67.1){\tiny \textbf{\underline{NC4K-\texttt{Te}~\cite{lyu2021simultaneously}}}}
    \put(50.5, 67.1){\tiny \textbf{\underline{CAMO-\texttt{Te}~\cite{le2019anabranch}}}}
    \put(81, 67.1){\tiny \textbf{\underline{COD10K-\texttt{Te}~\cite{fan2021concealed}}}}
    \put(20, 98.8){\tiny \textbf{\underline{NC4K-\texttt{Te}~\cite{lyu2021simultaneously}}}}
    \put(50.5, 98.8){\tiny \textbf{\underline{CAMO-\texttt{Te}~\cite{le2019anabranch}}}}
    \put(81, 98.8){\tiny \textbf{\underline{COD10K-\texttt{Te}~\cite{fan2021concealed}}}}
    \end{overpic}
    \caption{\jgp{The PR curves (1$^{st}$ row), F-measure curves (2$^{nd}$ row), and E-measure curves (3$^{rd}$ row) of COD-related competitors on three popular datasets. The closer the PR curve is to the upper-right corner, the better the performance is. The higher the F-measure/E-measure curve is, the better the performance the better the model works. Best viewed in color.}}
    \label{fig:pr}
    \vspace{-5pt}
\end{figure*}

\begin{figure*}[t!]
    \centering
    \begin{overpic}[width=0.97\linewidth]{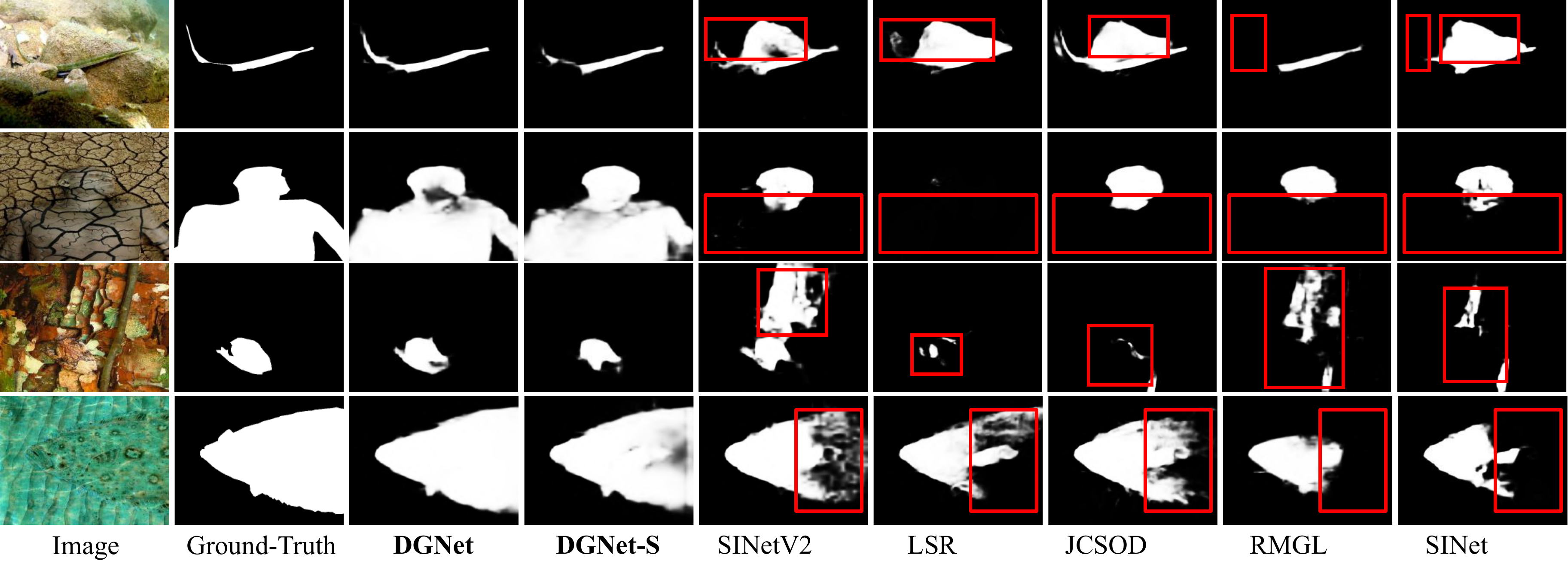}
    \put(52, 1.2){\footnotesize \cite{fan2021concealed}}
    \put(61.2, 1.2){\footnotesize \cite{lyu2021simultaneously}}
    \put(73.4, 1.2){\footnotesize \cite{li2021uncertainty}}
    \put(84.8, 1.2){\footnotesize \cite{zhai2021mutual}}
    \put(95.2, 1.2){\footnotesize \cite{fan2020camouflaged}}
    \end{overpic}
    \caption{Visualization of popular COD baselines and the proposed \ourmodel. Red boxes denote false-positive\res{/-negative} predictions. More results are presented at \href{https://github.com/GewelsJI/DGNet}{https://github.com/GewelsJI/DGNet}.
    }
    \label{fig:qulitative_result}
\end{figure*}

\begin{figure*}[t!]
    \centering
    \begin{overpic}[width=0.99\linewidth]{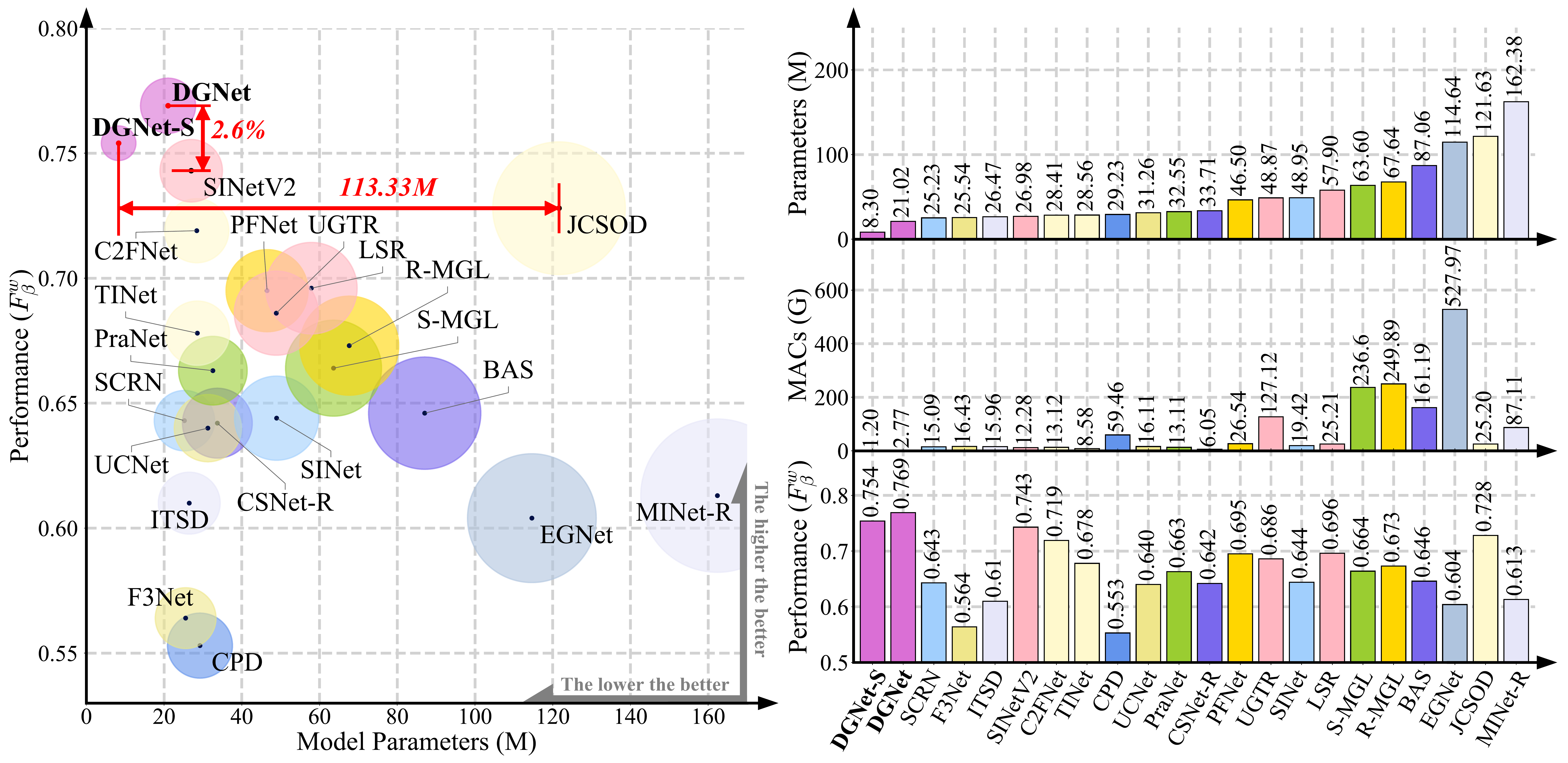}
    \put(36, 45){\tiny \underline{\textbf{CAMO-\texttt{Te}~\cite{le2019anabranch}}}}
    \end{overpic}
    \caption{(\textbf{Left}) We present the scatter relationship between the performance ($F_\beta^w$) and parameters of all competitors on CAMO-\texttt{Te}~\cite{le2019anabranch}. The larger the colored scatter point size, the heavier the model parameters. (\textbf{Right}) We also report the parallel histogram comparison of model's parameters, MACs, and performance ($F_{\beta}^{w}$). Best viewed in color.
    }
    \label{fig:efficiency_comp}
\end{figure*}

\subsection{Results and Analysis}

\noindent\textbf{Quantitative Results.}
As shown in~\tabref{tab:fullmetric_rank}, \ourmodel~achieves the \jgp{promising} performance in terms of all metrics. 
Especially, the gradient-based learning strategy helps to improve the completeness of predictions, providing a 2.6\% gain of $F_\beta^w$ on CAMO-\texttt{Te} than rank@$1$ model SINetV2~\cite{fan2021concealed}.

\noindent\jgp{\textbf{Quantitative Curves.}
As shown in \figref{fig:pr}, we plot the precision-recall (1$^{st}$ row), F-measure (2$^{nd}$ row), and E-measure (3$^{rd}$ row) curves of all COD-related competitors via varying with different thresholds. 
All comparisons show that our curves with magenta solid/dotted lines are significantly better than other methods on three datasets.}

\noindent\textbf{Qualitative Results.}
The visual comparison of four top-tier COD baselines and our \ourmodel~are shown in~\figref{fig:qulitative_result}.
Interestingly, these competitors fail to provide complete segmentation results for the camouflaged objects touching the image boundary. By contrast, our approach can precisely locate the target region and provide exact predictions due to the gradient learning strategy. 

\noindent\textbf{Efficiency Analysis.}
To better unveil the trade-off, two instances consistently obtain the best trade-off compared to existing competitors (see \figref{fig:efficiency_comp}).
DGNet outperforms cutting-edge model SINetV2~\cite{fan2021concealed} with a large margin ($F_\beta^w$: $+2.6\%$).
Notably, our efficient instance \ourmodelS~performs better than JCSOD~\cite{li2021uncertainty}, with 113.33M fewer parameters.
Besides, we also report the runtime comparison of all COD-related competitors in~\tabref{tab:runtime}, which are tested on an NVIDIA RTX TITAN GPU.
It clearly illustrates that \ourmodelS~and~\ourmodelL~can achieve super real-time inference speed (\ie, 80 fps \& 58 fps).

\begin{table*}[t!]
    \centering
    \footnotesize
    \caption{Inference speed (fps) among 12 COD-related models and the proposed two instances (\ie, \ourmodelS~and~\ourmodelL).}
    \label{tab:runtime}
    \renewcommand{\arraystretch}{1.0}
    \renewcommand{\tabcolsep}{8.5pt}
    \begin{tabular}{c||ccccccc}
    \hline
    Model &\cellcolor{gray!9} \textbf{\ourmodelS}&\cellcolor{gray!9} \textbf{\ourmodelL}  
    &SINetV2~\cite{fan2021concealed} &JCSOD~\cite{li2021uncertainty} &LSR~\cite{lyu2021simultaneously} &R-MGL~\cite{zhai2021mutual} &S-MGL~\cite{zhai2021mutual} \\
    \hline
    Input Size & \cellcolor{gray!9}352$\times$352 & \cellcolor{gray!9}352$\times$352 &352$\times$352 &352$\times$352 &352$\times$352 &473$\times$473 &473$\times$473 \\
    \hline
    Speed (fps) & \cellcolor{gray!9}\textbf{80} & \cellcolor{gray!9}58 &68 &43 &31 &9 &13 \\
    \hline
    Model &PFNet~\cite{mei2021camouflaged} &UGTR~\cite{yang2021uncertainty} &TINet~\cite{zhu2021inferring} &C2FNet~\cite{sun2021Context} &BAS~\cite{qin2021boundary} &PraNet~\cite{fan2020pranet} &SINet~\cite{fan2020camouflaged} \\
    \hline
    Input Size &416$\times$416 &473$\times$473 &352$\times$352 &352$\times$352 &288$\times$288 &352$\times$352 &352$\times$352\\
    \hline
    Speed (fps) &78 &15 &50 &68 &31 &73 &63 \\
    \hline
    \end{tabular}
    \vspace{-10pt}
\end{table*}

\begin{table*}[t!]
    \centering
    \footnotesize
    \renewcommand{\arraystretch}{1.1}
    \setlength\tabcolsep{6.5pt}
    \caption{Ablation studies. \texttt{\#Para} and \texttt{\#MACs} denote the parameters and multiply-accumulate operations of the model.}
    \label{tab:ablation_1}
    \begin{tabular}{c|c||rr|ccc|ccc|ccc}
    \hline
    & &\multicolumn{2}{c|}{Efficiency}
    &\multicolumn{3}{c|}{NC4K-\texttt{Te}~\cite{lyu2021simultaneously}} &\multicolumn{3}{c|}{CAMO-\texttt{Te}~\cite{le2019anabranch}} &\multicolumn{3}{c}{COD10K-\texttt{Te}~\cite{fan2021concealed}} \\
    No.& Variant &\texttt{\#Para} &\texttt{\#MACs}
    & $\mathcal{S}_{\alpha}\uparrow$ &$F_\beta^w\uparrow$ & $\mathcal{M}\downarrow$
    & $\mathcal{S}_{\alpha}\uparrow$ &$F_\beta^w\uparrow$ & $\mathcal{M}\downarrow$ 
    & $\mathcal{S}_{\alpha}\uparrow$ &$F_\beta^w\uparrow$ & $\mathcal{M}\downarrow$ \\
    \hline
    \hline
    \rowcolor{mygray}
    $\#\textbf{S}$ &\textbf{\ourmodelS} &8.30M &1.20G 
    &.845 &.764 &.047
    &.826 &.754 &.063
    &.810 &.672 &.036 \\
    \hline
    \multicolumn{10}{l}{(a) Base Network $\rightarrowtail$ see~\secref{sec:context_path}} \\
    \hline
    $\#01$ &Base &8.24M &0.58G
    &.834 &.676 &.061
    &.814 &.670 &.072
    &.793 &.550 &.049 \\ 
    \hline
    \multicolumn{10}{l}{(b) Configuration of Dimensional Reduction $\rightarrowtail$ see~\secref{sec:context_path}} \\
    \hline
    $\#02$ &$C_i={16}$ &8.00M &0.81G
    &.842 &.758 &.048
    &.824 &.749 &.066
    &.806 &.663 &.037 \\
    $\#03$ &$C_i={64}$ &9.36M &2.69G
    &.845 &.764 &.047
    &.827 &.748 &.065
    &.812 &.673 &.036 \\
    $\#04$ &$C_i={128}$ &13.30M &8.55G
    &.847 &.768 &.046
    &.828 &.751 &.062
    &.810 &.672 &.036 \\
    \hline
    \multicolumn{10}{l}{(c) Network Decoupling Strategy $\rightarrowtail$ see~\secref{sec:texture_path}} \\
    \hline
    $\#05$ & \textit{w/}~$\mathbf{X}_2$ &8.24M & 0.59G
    &.840 &.712 &.055
    &.822 &.701 &.074
    &.805 &.597 &.043 \\
    \hline
    \multicolumn{10}{l}{(d) Should we use $\mathbf{Z}^G$ as supervision? $\rightarrowtail$ see~\equref{equ:grad_descriptor}} \\
    \hline
    $\#06$ & \textit{w/}~$\mathbf{Z}^B$ &8.30M &1.20G
    &.841 &.753 &.049
    &.821 &.737 &.067
    &.804 &.654 &.038\\
    \hline
    \multicolumn{10}{l}{(e) Group Number $M$ $\rightarrowtail$ see~\equref{equ:group_number}} \\
    \hline
    $\#07$ &$M=1$ &8.30M &1.20G
    &.841 &.756 &.049
    &.822 &.751 &.064
    &.806 &.662 &.037 \\
    $\#08$ &$M=4$ &8.30M &1.20G
    &.842 &.759 &.048
    &.822 &.742 &.067 
    &.809 &.669 &.036 \\
    $\#09$ &$M=16$ &8.30M &1.20G
    &.842 &.752 &.049
    &.829 &.744 &.065  
    &.803 &.651 &.039 \\
    $\#10$ &$M=32$ &8.30M &1.20G
    &.845 &.913 &.047
    &.827 &.745 &.063 
    &.809 &.666 &.036 \\
    \hline
    \multicolumn{10}{l}{(f) Scaling Factors $N \in \{N_1,N_2,N_3\}$ $\rightarrowtail$ see~\equref{equ:group_att}} \\
    \hline
    $\#11$ &$\{2,4,8\}$ &8.31M &1.20G
    &.842 &.755 &.048
    &.821 &.741 &.065 
    &.808 &.663 &.036 \\
    $\#12$ &$\{4,8,16\}$ &8.30M &1.20G
    &.844 &.762 &.047
    &.823 &.744 &.065
    &.806 &.666 &.037  \\
    \hline
    \multicolumn{10}{l}{\res{(g) More sub-branches in Soft Grouping Strategy $\rightarrowtail$ see~\equref{equ:group_att}}} \\
    \hline
    \res{$\#13$} &$N \in \{4,8,16,32\}$ &8.30M &1.20G
    &.844 &.760 &.048
    &.829 &.748 &.064
    &.811 &.669 &.037 \\
    \res{$\#14$} &$N \in \{2,4,8,16,32\}$ &8.31M &1.20G
    &.846 &.765 &.047
    &.825 &.750 &.063
    &.810 &.670 &.037 \\ 
    \hline
    \multicolumn{10}{l}{\res{(h)} Gradient-Induced Transition $\rightarrowtail$ see~\equref{equ:transition}} \\
    \hline
    \res{$\#15$} &\textit{w/o} $\mathcal{T}_i$ &8.31M &1.20G
    &.839 &.748 &.050 
    &.825 &.741 &.065
    &.802 &.649 &.039 \\
    \hline
    \end{tabular}
\end{table*}

\subsection{Ablation Study}\label{sec:ablation}
We further ablate the core modules to verify the effectiveness of each part and configuration. For ecological reasons, we select the \ourmodelS~as the base model in this section.

\noindent\textbf{Contribution of Base Network.}
In \tabref{tab:ablation_1}-(a), we remove the texture encoder and GIT from \ourmodelS~and term it as the base network ($\#01$).
Compared to it, our \ourmodelS~($\#\textbf{S}$) significantly improves the performance while slightly increasing the model parameters by 0.06M.

\noindent\textbf{Configuration of Dimensional Reduction.}
We change the channel $C_i$ to 16 ($\#02$), 32 ($\#\textbf{S}$), 64 ($\#03$), and 128 ($\#04$) and find that more parameters may lead to performance saturation. 
To achieve the best trade-off between resource and speed, we choose $C_i$$=$$32$ as the default setting.

\noindent\textbf{Contribution of Network Decoupling Strategy.}
We explore the necessity of our decoupling strategy.
Inspired by~\cite{ji2021fast}, we replace the feature extracted from the texture encoder with the low-level feature $\mathbf{X}_2$ from the context encoder, which yields a single-stream network ($\#05$).
Notably, we only change the extracting manner of texture features and preserve the gradient-wise supervision for both variants (\ie, $\#05$ \& $\#\textbf{S}$) to ensure unbiased ablation.
\tabref{tab:ablation_1}-(c) demonstrates that decoupling the network into two streams can improve the performance ($F_{\beta}^{w}$: +5.3\% on CAMO-\texttt{Te}), which benefits from the modelling of separated branches without feature ambiguity at different hierarchies.

\begin{figure}[t!]
    \centering
    \begin{overpic}[width=\linewidth]{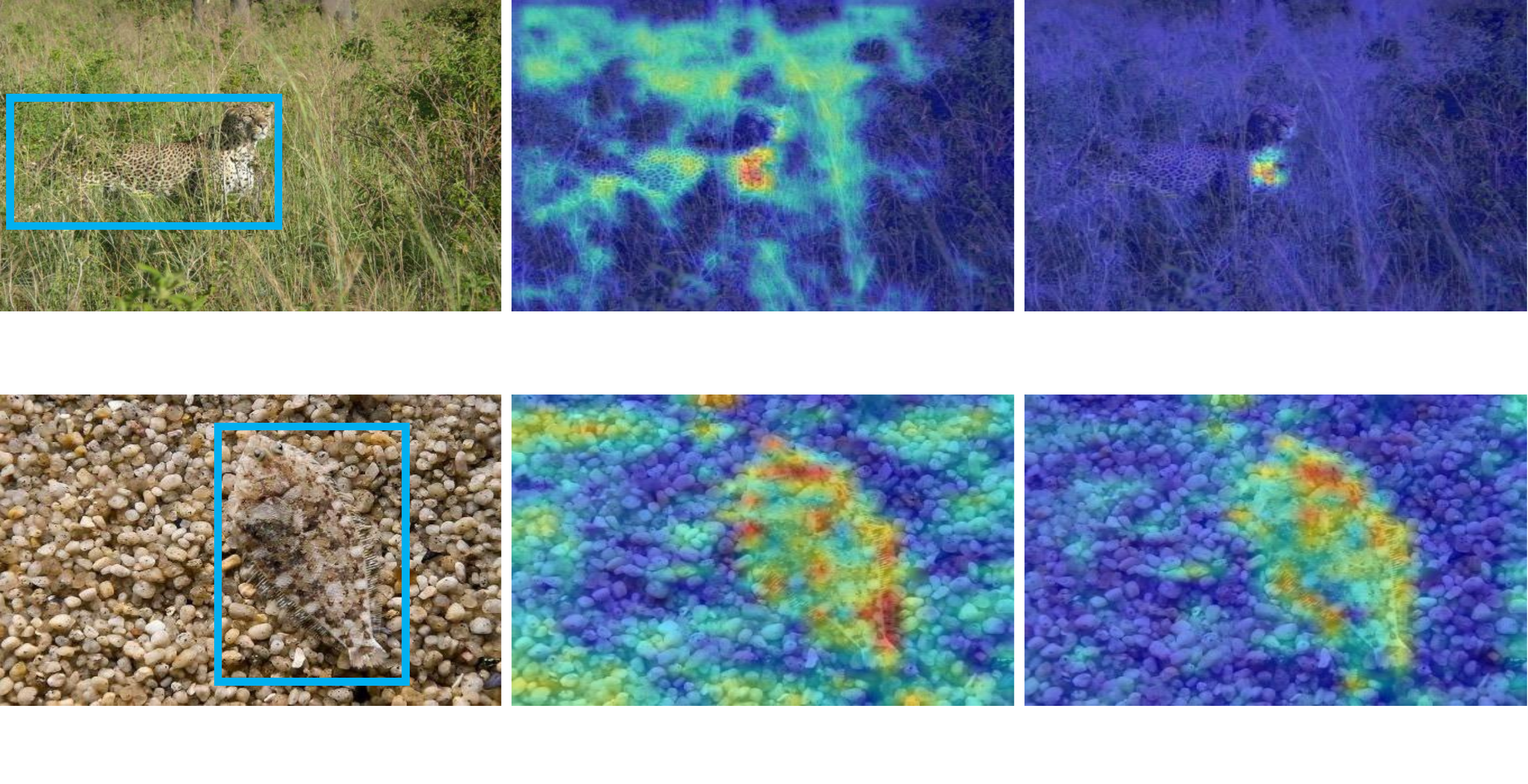}
    \put(0.5, 27.5){\footnotesize Camouflaged Image}
    \put(33, 27.5){\footnotesize (a) Object Boundary}
    \put(68, 27.5){\footnotesize (b) Object Gradient}
    \put(0.5, 2){\footnotesize Camouflaged Image}
    \put(37, 2){\footnotesize (c) Before GIT}
    \put(70, 2){\footnotesize (d) After GIT}
    \end{overpic}
    \caption{Feature visualizations of two core designs, including the supervision of object gradient (1$^{st}$ row) and the GIT (2$^{nd}$ row).}
    \label{fig:ablation_viz}
\end{figure}

\noindent\textbf{Contribution of Object Gradient Supervision.}
We replace the gradient map $\mathbf{Z}^G$ ($\#\textbf{S}$) with the boundary mask $\mathbf{Z}^B$ ($\#06$) \res{to supervise} the context learning process.
The improvement ($F_\beta^{w}$: $+1.7\%$ on CAMO-\texttt{Te}) of our gradient map supervision further demonstrates the effectiveness.
The first row of~\figref{fig:ablation_viz} presents the low-level features extracted from the texture learning branch under different supervision types. It shows that our solution can enforce the network to capture the gradient-sensitive information inside the camouflaged object's body, where those pixels learn to draw the observer's attention.

\begin{table}[t!]
    \footnotesize
    \renewcommand{\arraystretch}{1.2}
    \setlength\tabcolsep{0.3pt}
    \caption{Training \ourmodelS~under the supervision of texture label (TINet-Text~\cite{zhu2021inferring}) and our object gradient label (DGNet-Grad).}\label{tab:tinet_comp}
    \begin{tabular}{l|ccc|ccc|ccc}
    \hline
    & \multicolumn{3}{c|}{\tabincell{c}{NC4K-\texttt{Te}}}
    & \multicolumn{3}{c|}{\tabincell{c}{CAMO-\texttt{Te}}}
    & \multicolumn{3}{c}{\tabincell{c}{COD10K-\texttt{Te}}} \\ 
    & $\mathcal{S}_{\alpha}$ &$F_\beta^w$ & $\mathcal{M}$
    & $\mathcal{S}_{\alpha}$ &$F_\beta^w$ & $\mathcal{M}$
    & $\mathcal{S}_{\alpha}$ &$F_\beta^w$ & $\mathcal{M}$ \\
    \hline
    w/ TINet-Text 
    & .839 & .747 & .050
    & .820 & .731 & .068
    & .803 & .652 & .040 \\
    \rowcolor{mygray}
    \textbf{w/ DGNet-Grad}
    & \textbf{.845} & \textbf{.764} & \textbf{.047}
    & \textbf{.826} & \textbf{.754} & \textbf{.063}
    & \textbf{.810} & \textbf{.672} & \textbf{.036} \\
    \hline
    \end{tabular}
\end{table}

We further \res{experiment} using the supervision of texture labels~\cite{zhu2021inferring} (see~\figref{fig:grad_comp}), the results in \tabref{tab:tinet_comp} demonstrate that our gradient-supervision manner (\ie, w/ DGNet-Grad) is better than the texture-supervision (\ie, w/ TINet-Text). Besides, our method is simpler and more efficient than TINet, \eg, \ourmodelS~(8.0M) \textit{vs.} TINet (28.6M), \ourmodelS~(80 fps) \textit{vs.} TINet (50 fps). With such a compact design, we also achieve the new SOTA performance on CAMO-\texttt{Te}, \eg, \ourmodelS~($\mathcal{S}_\alpha=0.826$),  \ourmodel~($\mathcal{S}_\alpha=0.839$) \textit{vs.} TINet ($\mathcal{S}_\alpha=0.781$).

\noindent\textbf{Configuration of Group Numbers.}
In~\tabref{tab:ablation_1}-(e), we report different variants with respect to four group numbers $M$, which are equal to 1 ($\#07$), 4 ($\#08$), 8 ($\#\textbf{S}$), 16 ($\#09$), and 32 ($\#10$), respectively.
Note that $\#07$ ($M=1$) means ungrouped candidate features, which causes degraded performance ($F_\beta^{w}$: $-1.0\%$ on COD10K-\texttt{Te}).
We empirically choose $M=8$ with the best performance.

\noindent\textbf{Configuration of Scaling Factors.}
We also discuss how scaling factors $N \in \{N_1,N_2,N_3\}$ affect the model performance in \tabref{tab:ablation_1}-(f).
Compared with different configurations ($\#11$: $\{2,4,8\}$ and $\#12$: $\{4,8,16\}$), our finer-grained factors ($\#\textbf{S}$: $\{8,16,32\}$) lead to better prediction performance.
As shown in~\figref{fig:git}, we present the feature visualization of three parallel features (\ie, $\mathbf{A}^{N_1}_i$, $\mathbf{A}^{N_2}_i$, and $\mathbf{A}^{N_3}_i$), where the network puts different attention weights on different parts of the object insides.
This also validates that parallel residual learning can enhance the context feature from different group-aware perspectives.

\noindent\textbf{Do we need more sub-branches for soft grouping?} 
\res{As shown in~\tabref{tab:ablation_1}-(g), we set three ablative experiments for different sub-branches: three  ($\#\textbf{S}$: $N \in \{8,16,32\}$), four ($\#13$: $N \in \{4,8,16,32\}$), and five ($\#14$: $N \in \{2,4,8,16,32\}$) sub-branches. The comparison results unveil that more sub-branches would present unstable performance on all the datasets.}

\noindent\textbf{Contribution of Gradient-Induced Transition.}
We replace the whole GIT in our model with the naive channel-wise concatenation (\res{$\#15$: \textit{w/o}~$\mathcal{T}_i$ in ~\tabref{tab:ablation_1}-(h)}) to verify its effectiveness, which shows that our \ourmodelS~equipped with GIT ($\#\textbf{S}$: \textit{w/}~$\mathcal{T}_i$) can improve 2.3\% $F^{w}_\beta$ on the COD10K-\texttt{Te} dataset.
Moreover, as shown in the second-row of~\figref{fig:ablation_viz}, the model obtains a cleaner and finer representation $\mathbf{Z}^T_i$ (\ie, (d) after GIT) while suppressing the noises in the background of $\mathbf{X}^R_i$ (\ie, (c) before GIT).
A clear benefit of the adaptive aggregation of the context and texture cues in the GIT.

\subsection{Limitations}

\begin{table}[t!]
    \footnotesize
    \renewcommand{\arraystretch}{1.2}
    \setlength\tabcolsep{1.6pt}
    \caption{\res{Our method with different backbones, including EfficientNet~\cite{tan2019efficientnet} (\ie, EffNet-B1 \& EffNet-B4) \textit{vs.} MobileNet~\cite{Howard_2019_ICCV} (\ie, MobNet-S \& MobNet-L).}}\label{tab:mobilenet}
    \begin{tabular}{l|rr|cc|cc|cc}
    \hline
    &&& \multicolumn{2}{c|}{\tabincell{c}{NC4K-\texttt{Te}}}
    & \multicolumn{2}{c|}{\tabincell{c}{CAMO-\texttt{Te}}}
    & \multicolumn{2}{c}{\tabincell{c}{COD10K-\texttt{Te}}} \\
    &\texttt{\#Para} &\texttt{MACs} & $\mathcal{S}_{\alpha}$ &$F_\beta^{w}$
    & $\mathcal{S}_{\alpha}$ &$F_\beta^{w}$
    & $\mathcal{S}_{\alpha}$ &$F_\beta^{w}$\\
    \hline
    MobNet-S   
    & 2.96M & 1.27G & .779 & .638  
    & .735 & .587 
    & .729 & .517 \\
    \rowcolor{mygray}
    \textbf{EffNet-B1}
    & 8.30M & 1.20G & .845 & .764
    & .826 & .754
    & .810 & .672 \\
    MobNet-L
    & 6.96M & 3.17G & .820 & .723
    & .791 & .686
    & .780 & .620 \\
    \rowcolor{mygray}
    \textbf{EffNet-B4}
    & 21.02M & 2.77G & .857 & .784
    & .839 & .769
    & .822 & .693\\
    \hline
    \end{tabular}
\end{table}

\noindent\textbf{Efficient Backbone \textbf{vs.} Lightweight one.}
\res{We further validate the potential value of our method on limited hardware conditions by replacing the efficient backbone, EfficientNet~\cite{tan2019efficientnet}, with a lightweight one, MobileNet~\cite{Howard_2019_ICCV}. The results, as in~\tabref{tab:mobilenet}, show that our method achieves unsatisfactory performance with a lightweight backbone, \ie, MobNet-S (2.96M) and MobNet-L (6.96M), leaving a huge room for our future exploration.}

\begin{figure}[t!]
    \centering
    \begin{overpic}[width=\linewidth]{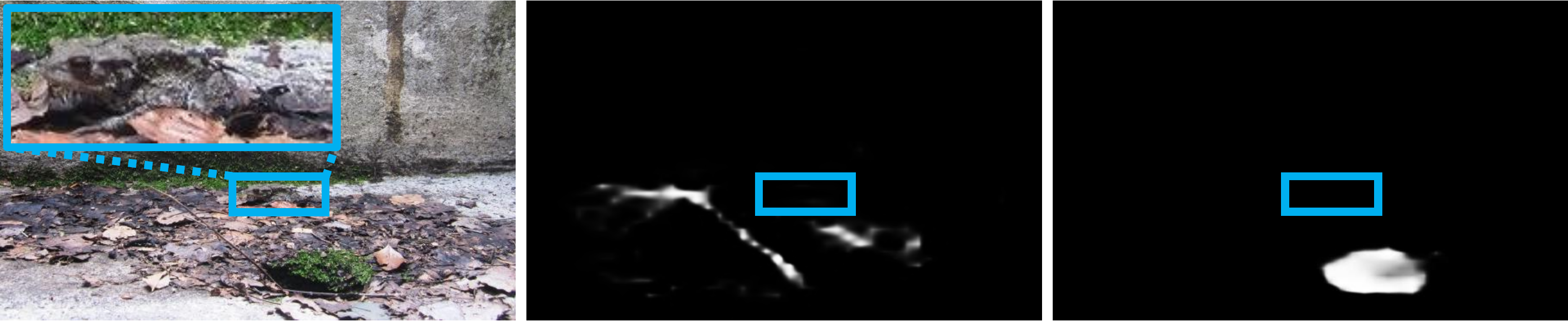}
    \put(0, -4){\footnotesize Camouflaged Image}
    \put(40, -4){\footnotesize SINetV2~\cite{fan2021concealed}}
    \put(69, -4){\footnotesize \textbf{\ourmodel~(Ours)}}
    \end{overpic}
    \vspace{1pt}
    \caption{Hard sample with small camouflaged object.}
    \label{fig:failure_case}
\end{figure}

\noindent\textbf{Challenging Cases.}
Despite our method's satisfactory performance, it may fail in challenging camouflaged scenarios \res{as follows.
First,} we argue that in the proposed strategy it is still difficult to provide enough texture cues in the limited small target region, resulting in false-positive predictions.
As shown in~\figref{fig:failure_case}, such cases also easily confuse the rank@$1$ approach SINetV2~\cite{fan2021concealed}, thus deserving further studies.

\begin{figure}[t!]
    \centering
    \includegraphics[width=\linewidth]{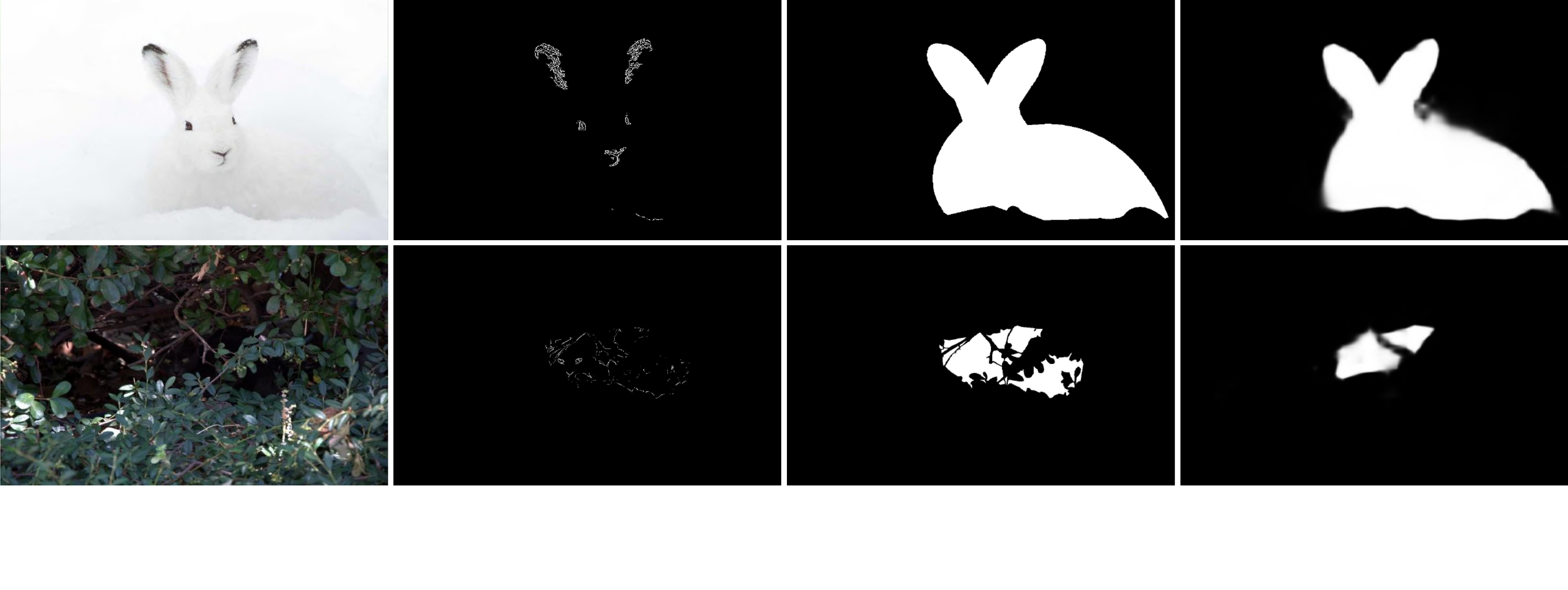}
    \put(-212, 7){\footnotesize Camouflaged}
    \put(-200, -1){\footnotesize Image}
    \put(-146, 7){\footnotesize Object}
    \put(-150, -1){\footnotesize Gradient}
    \put(-105, 7){\footnotesize Ground-Truth}
    \put(-45, 7){\footnotesize Prediction}
    \vspace{1pt}
    \caption{\res{Visual comparison of the object with non-distinct object gradient cues.}}\label{fig:limitation_gradient_cues}
\end{figure}
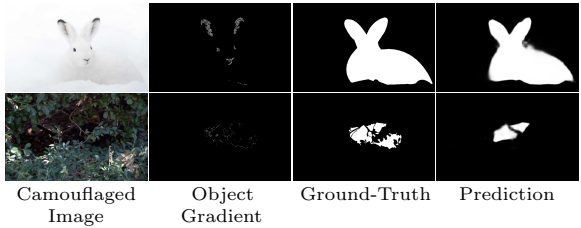

\res{Second, we observe that not all the camouflaged objects with noticeable gradient changes inside themselves. As shown in the first row of \figref{fig:limitation_gradient_cues}, our method could segment a white rabbit with non-distinct gradient changes. 
However, our method fails under extreme conditions, as in the second row of \figref{fig:limitation_gradient_cues}, which has rare gradient cues. It needs to design by incorporating more heuristic and learnable patterns for future improvements.}

\res{Additionally, we noticed a recently released COD method, ZoomNet~\cite{pang2022zoom}, after the submission. As shown in~\tabref{tab:zoomnet}, our \ourmodel~surpass the ZoomNet with a margin (\ie, NC4K-\texttt{Te}: +1.3\% $E_\phi^{mx}$ and CAMO-\texttt{Te}: +2.3\% $E_\phi^{mx}$), but fails to outperform ZoomNet on COD10K-\texttt{Te}. ZoomNet occupies more computational costs (32.38M parameters) than our DGNet (21.02M parameters). It inspires us to incorporate the zooming strategy into our network for our future extension.}

\begin{table}[t!]
    \footnotesize
    \renewcommand{\arraystretch}{1.2}
    \setlength\tabcolsep{1.3pt}
    \caption{\res{Performance comparison of our \ourmodel~and the recent released ZoomNet~\cite{pang2022zoom} on three testing datasets.}}\label{tab:zoomnet}
    \begin{tabular}{l|rr|cc|cc|cc}
    \hline
    &&& \multicolumn{2}{c|}{\tabincell{c}{NC4K-\texttt{Te}}}
    & \multicolumn{2}{c|}{\tabincell{c}{CAMO-\texttt{Te}}}
    & \multicolumn{2}{c}{\tabincell{c}{COD10K-\texttt{Te}}} \\
    &\texttt{\#Para} &\texttt{MACs} & $\mathcal{S}_{\alpha}$ &$E_\phi^{mx}$
    & $\mathcal{S}_{\alpha}$ &$E_\phi^{mx}$
    & $\mathcal{S}_{\alpha}$ &$E_\phi^{mx}$\\
    \hline
    ZoomNet 
    &32.38M &34.96G & .853 & .912
    & .820 & .892
    & \textbf{.838} & \textbf{.911}\\
    \rowcolor{mygray}
    \textbf{DGNet}
    &\textbf{21.02M} &\textbf{2.77G} & \textbf{.857} & \textbf{.922}
    & \textbf{.839} & \textbf{.915}
    & .822 & \textbf{.911}\\
    \hline
    \end{tabular}
\end{table}

\section{Downstream Applications}\label{sec:apps}
This section also assesses the generalization capabilities of three downstream applications.

\begin{figure*}[t!]
    \includegraphics[width=\linewidth]{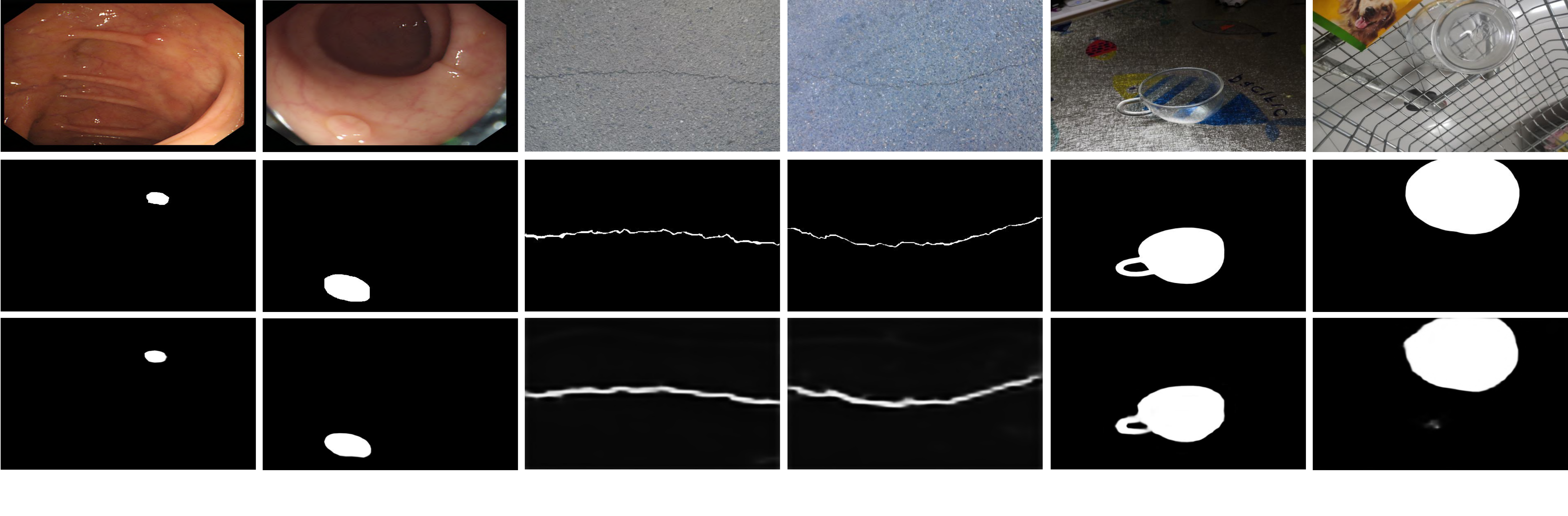}
    \put(-420, 1){\footnotesize (a) Polyp Segmentation}
    \put(-275, 1){\footnotesize (b) Road Crack Detection}
    \put(-140, 1){\footnotesize (c) Transparent Object Segmentation}
    \caption{Visualization results of three downstream applications. From top to bottom: input image (1$^{st}$ row), ground-truth (2$^{nd}$ row), and prediction (3$^{rd}$ row).}
\label{fig:applications}
\end{figure*}

\noindent\textbf{Polyp Segmentation.}
In the early diagnosis of colonoscopy, the low boundary contrast between a polyp and its highly-similar surroundings significantly decreases the detectability of colorectal cancer.
To demonstrate the generality of our method in the medical field, we follow the same benchmark protocols as~\cite{fan2020pranet} and retrain our \ourmodel~on the training set of Kvasir-SEG~\cite{jha2020kvasir} and CVC-ClinicDB~\cite{bernal2015wm} datasets.
We use two unseen test datasets: CVC-ColonDB~\cite{bernal2012towards} and ETIS-LPDB~\cite{silva2014toward}.
\tabref{tab:application_1} shows that our $^\dag$\ourmodel consistently surpasses four cutting-edge polyp segmentation methods in four metrics, including $\mathcal{S}_\alpha$, $E^{mx}_\phi$, $F^{w}_\beta$, and maximum Dice score ($D^{mx}$). Notably, $^\dag$\ourmodel~denotes that we retrain \ourmodel~on the task-specific training dataset.
\figref{fig:applications}~(a) shows the visualization results generated by our $^\dag$\ourmodel.

\begin{table}[t!]
    \centering
    \footnotesize
    \renewcommand{\arraystretch}{1.0}
    \setlength\tabcolsep{0pt}
    \caption{Quantitative results on two popular polyp segmentation test datasets.}
    \label{tab:application_1}
    \begin{tabular}{r|cccc|cccc}
    \hline
    &\multicolumn{4}{c|}{CVC-ColonDB~\cite{bernal2012towards}} &\multicolumn{4}{c}{ETIS-LPDB~\cite{silva2014toward}} \\
    Baseline~
    & $\mathcal{S}_{\alpha}\uparrow$ &$E^{mx}_\phi\uparrow$ & $F^w_\beta\uparrow$ &$D^{mx}\uparrow$ 
    & $\mathcal{S}_{\alpha}\uparrow$ &$E^{mx}_\phi\uparrow$ & $F^w_\beta\uparrow$ &$D^{mx}\uparrow$  \\
    \hline
    UNet~\cite{ronneberger2015u}
    &.710&.781&.491&.560
    &.684&.740&.366&.444
    \\
    UNet++~\cite{zhou2019unet++}
    &.692&.764&.467&.550
    &.683&.776&.390&.509
    \\
    PraNet~\cite{fan2020pranet}
    &.820&.872&.699&.728
    &.794&.841&.600&.639
    \\
    MSNet~\cite{zhao2021automatic}
    &.838&.883&.736&.766
    &.845&.890&.677&.736
    \\
    \hline
    \rowcolor{mygray}
    \textbf{$^\dag$\ourmodel}
    &\textbf{.858}&\textbf{.898}&\textbf{.765}&\textbf{.789}
    &\textbf{.847}&\textbf{.904}&\textbf{.690}&\textbf{.741}
    \\
    \hline
    \end{tabular}
\end{table}

\noindent\textbf{Defect Detection.}
Substandard products (\eg, tiles, wood) will inevitably incur unrecoverable economic losses in manufacturing.
We further retrain our $^\dag$\ourmodel~on the road crack detection dataset (\ie, CrackForest~\cite{shi2016automatic}), using 60\% of the samples for training and 40\% for testing.
\figref{fig:applications}~(b) presents some visualization cases.

\noindent\textbf{Transparent Object Segmentation.}
In daily life, intelligent agents such as robots and drones need to identify unnoticeable transparent objects (\eg, glasses, bottles, and mirrors) to avoid accidents.
We also verify the effectiveness of the retrained model $^\dag$\ourmodel~on the transparent object segmentation task.
For convenience, we re-organize the annotation of the Trans10K~\cite{xie2020segmenting} dataset from instance-level to object-level for training. 
The visual results shown in~\figref{fig:applications}~(c) further demonstrate the learning ability of $^\dag$\ourmodel.

\section{Conclusion}\label{sec:Conclusion}
We presented a novel deep gradient learning framework (\textbf{\ourmodel}) for efficiently segmenting camouflaged objects.
To extract the camouflaged features, we proposed to decouple the task into two branches, a context encoder and a texture encoder. 
We designed a novel plug-and-play module called gradient-induced transition (GIT), acting as a soft grouping module to learn features from these two branches jointly. 
This simple and flexible architecture showed strong generalization capabilities on three challenging datasets compared to the 20 SOTA competitors.
In addition, our efficient version \textbf{\ourmodelS}~(8.3M \& 80 fps) achieved an excellent performance-efficiency trade-off.
Our solution also produced visually appealing results for three further applications, including polyp segmentation, defect detection, and transparent object segmentation, which validates its practical application value.

\bibliographystyle{IEEEtran}
\bibliography{mir-article}

\end{document}